\documentclass[10pt,twocolumn,letterpaper]{article}

\usepackage{config/style}

\def\cvprPaperID{*****}
\def\confName{CVPR}
\def\confYear{2023}

\usepackage[pagebackref=true,breaklinks=true,colorlinks=true,citecolor=citecolor,urlcolor=magenta,bookmarks=true]{hyperref}
\usepackage{amsmath}
\usepackage[capitalize]{cleveref}
\usepackage{lipsum}
\usepackage[dvipsnames]{xcolor}
\usepackage{color, colortbl}
\usepackage{amsthm}
\usepackage{wrapfig}
\usepackage{array}
\usepackage{newlfont}
\usepackage{textcomp}
\usepackage{multirow}
\usepackage{commath}
\usepackage{hhline} 
\usepackage{bigstrut}
\usepackage{afterpage}
\usepackage{algorithmic}
\usepackage{microtype}
\usepackage{mathtools}
\usepackage{booktabs}
\usepackage[ruled]{algorithm2e}
\usepackage{cite}
\usepackage{dingbat}
\usepackage{balance}
\usepackage{tabularx}

\usepackage{times}
\usepackage{epsfig}
\usepackage[export]{adjustbox}
\usepackage{graphicx}
\usepackage{amssymb}
\usepackage{cuted}
\usepackage{capt-of}
\usepackage{float}
\usepackage{enumitem}
\usepackage{comment}
\usepackage{pifont}
\usepackage[toc,page]{appendix}
\usepackage{soul}
\usepackage{acronym}
\usepackage[usestackEOL]{stackengine}
\usepackage{fancybox}

\definecolor{citecolor}{HTML}{0071bc}
\definecolor{frontcolor}{HTML}{325ea5}
\definecolor{sidecolor}{HTML}{a58b77}
\definecolor{DeltaColor}{rgb}{0.039,0.73,0.71}
\definecolor{SigmaColor}{rgb}{0.98,0.45,0.0}
\definecolor{AlphaColor}{rgb}{0,0,0.8}
\definecolor{BetaColor}{rgb}{0.8,0,0.8}
\definecolor{GammaColor}{rgb}{0.514,0.34,0.224}
\definecolor{EpsilonColor}{rgb}{0.353,0.725,0.906}
\definecolor{PurpleColor}{HTML}{bca5ea}
\definecolor{OrangeColor}{rgb}{0.914,0.541,0.0.141}
\definecolor{GreenColor}{rgb}{0.137,0.573,0.565}
\definecolor{RedColor}{rgb}{0.949,0.275, 0.224}
\definecolor{LightCyan}{rgb}{0.88,1,1}
\definecolor{Gray}{gray}{0.85}

\newcolumntype{a}{>{\columncolor{Gray}}c}

\newcommand{\colorRef}[1]{\textcolor{black}{#1}}

\crefname{figure}{\colorRef{Fig.}}{\colorRef{Figs.}}
\Crefname{figure}{\colorRef{Figure}}{\colorRef{Figures}}
\crefname{section}{\colorRef{Sec.}}{\colorRef{Secs.}}
\Crefname{section}{\colorRef{Section}}{\colorRef{Sections}}
\Crefname{table}{\colorRef{Table}}{\colorRef{Tables}}
\crefname{table}{\colorRef{Tab.}}{\colorRef{Tabs.}}

\newcolumntype{x}[1]{>{\centering\arraybackslash}p{#1pt}}
\newcolumntype{y}[1]{>{\raggedright\arraybackslash}p{#1pt}}
\newcolumntype{z}[1]{>{\raggedleft\arraybackslash}p{#1pt}}
\newlength\savewidth

\newcommand{\modelnameLong}{High-Fidelity Clothed Avatar Reconstruction from a Single Image\xspace}
\newcommand{\ourtitle}{ \modelnameLong}

\acrodef{amt}[AMT]{Amazon Mechanical Turk}

\makeatletter
\newcommand*{\addFileDependency}[1]{
  \typeout{(#1)}
  \@addtofilelist{#1}
  \IfFileExists{#1}{}{\typeout{No file #1.}}
}

\makeatother

\makeatletter
\newcolumntype{R}[2]{%
    >{\adjustbox{angle=#1,lap=\width-(#2)}\bgroup}%
    l%
    <{\egroup}%
}

\newcommand{\thickhline}{%
    \noalign {\ifnum 0=`}\fi \hrule height 1.5pt
    \futurelet \reserved@a \@xhline
} 
\newcolumntype{"}{@{\hskip\tabcolsep\vrule width 2pt\hskip\tabcolsep}}
\makeatother

\newcolumntype{I}{!{\vrule width 1pt}}
\newcommand*{\Hline}[0]{%
\noalign{\global\setlength{\arrayrulewidth}{1pt}}%
\hline
\noalign{\global\setlength{\arrayrulewidth}{0.4pt}}%
}
\newcommand*{\Cline}[1]{%
\noalign{\global\setlength{\arrayrulewidth}{1pt}}%
\cline{#1}%
\noalign{\global\setlength{\arrayrulewidth}{0.4pt}}%
}

\begin{document}

\title{\ourtitle}


\author{ 
    Tingting Liao$^{1,2}$\thanks{Equal contribution.} \qquad
    Xiaomei Zhang$^{1,2 *}$  \qquad
    Yuliang Xiu$^{3}$ \qquad
    Hongwei Yi$^{3}$ \qquad
    Xudong Liu$^{4}$ \qquad \\ 
    Guo-Jun Qi$^{4,5}$ \qquad
    Yong Zhang$^{6}$ \qquad
    Xuan Wang$^{6}$ \qquad
    Xiangyu Zhu$^{1,2}$ \qquad
    Zhen Lei$^{1,2,7}$\thanks{Corresponding author.}   \\
    {\normalsize $^{1}$University of Chinese Academy of Sciences, Beijing, China ~~~}    \\
    {\normalsize$^{2}$MAIS, Institute of Automation, Chinese Academy of Sciences, Beijing, China} \\
    {\normalsize$^{3}$Max Planck Institute for Intelligent Systems, T\"ubingen, Germany ~~~} \\
    {\normalsize$^{4}$OPPO Research  ~~~ 
    $^{5}$Westlake University  ~~~ 
    $^{6}$Tencent AI Lab ~~~ 
    $^{7}$CAIR, HKISI, CAS }\\ 
    {\tt\small \{tingting.liao, xiaomei.zhang, xiangyu.zhu, zlei\}@nlpr.ia.ac.cn} \\  
     {\tt\small \{yuliang.xiu, hongwei.yi\}@tuebingen.mpg.de} \\  
    {\tt\small  \{yongzhang201303, xwang.cv, guojunq\}@gmail.com} \\  
    {\tt\small  \{xudong.liu\}@oppo.com} \\ 
}

\newcommand{\teaserCaption}{
{\bf Images to avatars.}
Given an image of a person in an unconstrained pose (a) our method reconstructs 3D clothed avatars in both original posed space (b) and canonical space (c) and can repose the human body from the canonical mesh (d).
}

\twocolumn[{
    \renewcommand\twocolumn[1][]{#1}
    \maketitle
    \centering
    \vspace{-0.5em}
    \begin{minipage}{1.00\textwidth}
        \centering
        \includegraphics[trim=000mm 000mm 000mm 000mm, clip=False, width=\linewidth]{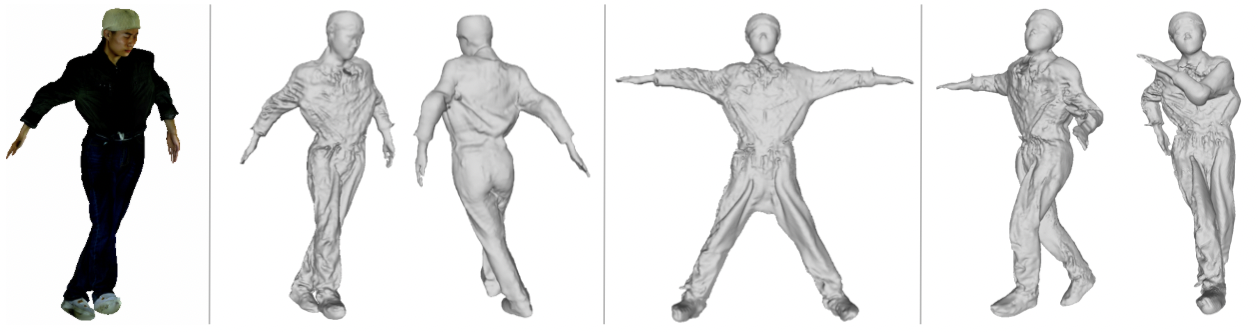}
    \rightline { \small   (a) Input \quad\quad\quad\quad\quad\quad  (b) Posed Reconstruction \quad\quad (c) Canonical Reconstruction \quad\quad\quad\quad\quad\quad  (d) Reposed \quad\quad\quad\quad\quad}  
    \end{minipage}
    \vspace{-1.0 em}
    \captionof{figure}{\teaserCaption}
    \label{fig:teaser}
    \vspace{1.2em}
}]

\begin{abstract}
This paper presents a framework for efficient 3D clothed avatar reconstruction. By combining the advantages of the high accuracy of optimization-based methods and the efficiency of learning-based methods, we propose a coarse-to-fine way to realize a high-fidelity clothed avatar reconstruction (CAR) from a single image. At the first stage, we use an implicit model to learn the general shape in the canonical space of a person in a learning-based way, and at the second stage, we refine the surface detail by estimating the non-rigid deformation in the posed space in an optimization way. A hyper-network is utilized to generate a good initialization so that the convergence o f the optimization process is greatly accelerated. Extensive experiments on various datasets show that the proposed CAR successfully produces high-fidelity avatars for arbitrarily clothed humans in real scenes. The codes will be released in \href{https://github.com/TingtingLiao/CAR}{https://github.com/TingtingLiao/CAR}. 
\end{abstract}
\section{Introduction}
\label{sec:intro} 

Clothed avatar reconstruction is critical to a variety of applications for 3D content creations such as video gaming, online meeting\cite{yi2022mime,yi2022generating}, virtual try-on and movie industry~\cite{bhatnagar2019multi,santesteban2019learning,hu3DBodyNet}. Early attempts are based on expensive scanning devices such as 3D and 4D scanners, or complicated multi-camera studios with carefully capturing processes. While highly accurate reconstruction results can be obtained from these recording equipment, they are inflexible and even not feasible in many applications. An alternative is to collect data using depth sensors~\cite{slavcheva2017killingfusion,newcombe2015dynamicfusion}, which is however still less ubiquitous than RGB cameras. A more practical and low-cost way is to create an avatar from an image by RGB cameras or mobile phones.
 
Monocular RGB reconstruction \cite{he2020geopifu, saito2020pifuhd, Zheng2019DeepHuman, xiu2022icon} has been extensively investigated and shows promising results. ARCH~\cite{huang2020arch} is the first method that reconstructs a clothed avatar from a monocular image. Due to the disadvantage of depth ambiguity, a number of methods that create an avatar from a video are proposed to resolve the problem. Most existing monocular video-based methods \cite{alldieck2018video, alldieck2018detailed, balan2007detailed, gall2009motion, vlasic2008articulated, furukawa2006carved, bhatnagar2019multi, bhatnagar2020loopreg} are typically restricted to parametric human body prediction, which lacks geometry details like cloth surface. How to create a high-fidelity avatar from an in-the-wild image, with consistent surface details is still a great challenge.

In this work, we focus on the shape recovery and propose an efficient high-fidelity clothed human avatar creation method from a single image in a coarse-to-fine way. The method consists of a learning-based canonical implicit model and an optimization-based normal refinement process. The canonical implicit model uses the canonical normal inverse transformed from original space as geometric feature to help grasp clothing detail of the general shape in canonical space. Unlike occupancy-based methods~\cite{saito2019pIFu, saito2020pifuhd, huang2020arch}, we adopt a Signed Distance Function (SDF) to approximate the canonical human body surface, which gains advantages in learning the human body in the surface level instead of the point level, so that the reconstruction accuracy is improved. In the normal refinement process, a SDF is learned to approximates the target surface in the posed space by enforcing its surface normal closed to the predicted normal image. Compared with mesh-based refinement, our method can obtain more realistic results without artifacts due to the flexibility of implicit representation. Moreover, to learn the SDF of the normal refinement process efficiently, we propose a meta-learning-based hyper network for parameter initialization to accelerate the convergence of the normal refinement process. 
 
Extensive experiments have been conducted on MVP-Human~\cite{zhu2022mvp} and RenderPeople~\cite{RenderPeople} datasets. Both qualitative and quantitative results demonstrate that our proposed method outperforms related avatar reconstruction methods. The main contributions are summarized as follows: 
 \begin{itemize} 
\item We propose a coarse-to-fine framework for efficient clothed avatar reconstruction from a single image. Thanks to the integration of image and geometry features, as well as the meta-learning, it achieves high-fidelity clothed avatar reconstruction efficiently.  
\item We design the canonical implicit regression model and the normal refinement process. The former fuses all observations to the canonical space where the general shape of a person is depicted, and the latter learns pose dependent deformation. 
\item Results validate that our method could reconstruct high-quality 3D humans in both posed and canonical space from a single in-the-wild image. 
\end{itemize} 
 
\section{Related Work}
\label{sec:related} 
\noindent \textbf{3D Clothed Human Reconstruction.} 3D clothed human reconstruction \cite{jiang2020bcnet,li2020monocular,li2020monoportRTL,lazova2019360,tang2019neural,xiu2022icon,alldieck2022phorhum} from multi-view or even a single image has achieved great progress in recent years. Saito et al. \cite{saito2019pIFu} introduce Pixel-Aligned Implicit Function (PIFu), which formulates an implicit function using pixel-aligned image features and points the depth to obtain the human body's occupancy field for the first time. However, PIFu cannot preserve high-frequency details like cloth wrinkles and then the generated surfaces tend to be smooth. To address this issue, PIFuHD~\cite{saito2020pifuhd} proposes a multi-level framework to reconstruct high-fidelity clothed humans from high-resolution normal images. Despite impressive results, both PIFu and PIFuHD show poor robustness on in-the-wild images with out-of-distribution poses. Some works \cite{zheng2021pamir,he2020geopifu ,xiu2022icon,xiu2023econ,bhatnagar2020ipnet} try to tackle this problem by utilizing the human body prior to learn 3D semantic information. The combination of explicit parametric models and implicit representations enables the model to be more robust to out-of-distribution poses. For example, GeoPIFu~\cite{he2020geopifu} and PaMIR~\cite{zheng2021pamir} extract voxel-aligned semantic features from SMPL \cite{SMPL:2015} body to make the model more robust to pose variation. Recently, Xiu et al. \cite{xiu2022icon} find that these methods are sensitive to the global pose, due to their 3D convolutional encoders. ICON~\cite{xiu2022icon} uses local features including normal and signed distance to estimate the occupancy value of a query point. PHORHUM \cite{alldieck2022photorealistic} additionally estimates the albedo and global scene illumination, hence enabling relighting. While impressive results can be obtained from existing methods, such approaches reconstruct static 3D humans which cannot be animated. 
    
\noindent \textbf{Avatar Reconstruction.} Many works \cite{saito2021scanimate, chen2021snarf, deng2020nasa, wang2021metaavatar} use scanning devices to obtain 4D scans and fuse them into an animatable avatar. Similarly, human performance capture approaches \cite{habermann2019livecap,habermann2020deepcap, xu2018monoperfcap,yu2021function4d,li2022avatarcap} use a pre-scanned template and track per-frame shape deformations. Nevertheless, all these methods require expensive and unportable capture devices. In contrast, RGB monocular camera-based avatar reconstruction gains more popularity in recent years due to its low cost and convenience. The methods can be roughly categorized into optimization-based and learning-based ones.  

\begin{figure*}[t]
\centering
\includegraphics[width=\linewidth]{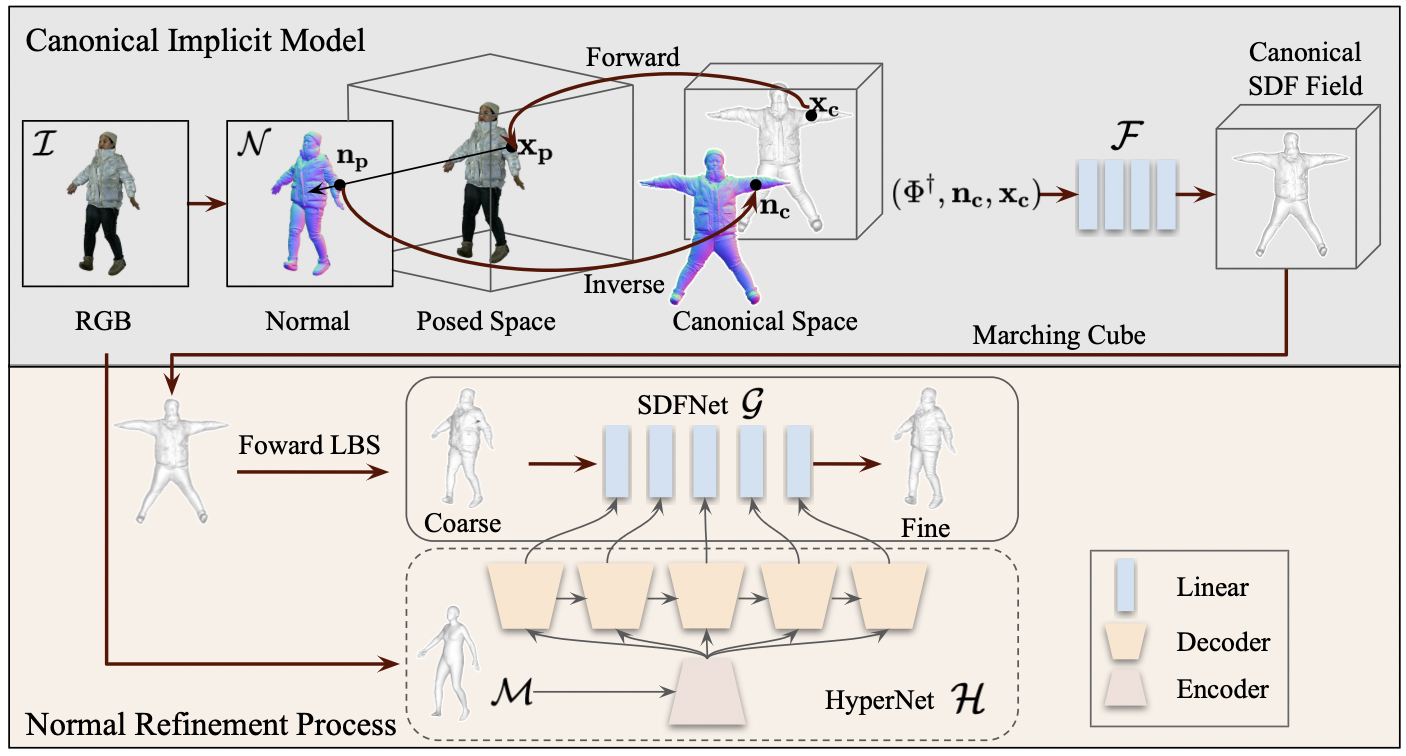}  
\vspace{-2.0 em}
\caption{Framework of CAR. Given an RGB image $\mathcal{I}$, we first estimate its SMPL body $\mathcal{M}$ and the normal map $\mathcal{N}$. The canonical implicit model then takes $\mathcal{N}$ and body pose as input to estimate a canonical SDF field. Then, the normal refinement process warps
it to the posed space and generates a high-fidelity clothed avatar reconstruction.}
\label{fig:overview}
\vspace*{-1. em} 
\end{figure*} 

The optimization-based methods focus on overfitting an avatar from a video of a specific moving subject. Early works \cite{esteban2004silhouette, furukawa2006carved, weng2019photo, starck2007surface} are based on visual hulls using silhouettes from multiple views to obtain the visible areas of the captured person. The concavity problem is changeable and difficult to handle. After that, researchers~\cite{alldieck2018video, balan2007detailed, alldieck2018detailed, santesteban2019learning, pons2017clothcap, xiang2020monoclothcap, bhatnagar2019multi} attempt to model the geometry on top of parametric models with vertex offsets. The mesh representation has a fixed topology which is insufficient to recover high-quality results, especially on loose clothes such as skirts and dresses. Unlike meshes, implicit
representations are more powerful to help recover detailed 3D shapes with
arbitrary topology. NeRF-based methods \cite{xu2021h,peng2021neural, zheng2022avatar} optimize a goal using conditions on articulated cues. Jiang et al.~\cite{jiang2022selfrecon} further combine the explicit and implicit representations to reconstruct high-fidelity geometry. However, the optimization process in these methods is time-consuming and inefficient in real applications. 

Comparatively, the learning-based methods are more efficient in the prediction process. ARCH \cite{huang2020arch} is the first work to propose an end-to-end learning-based framework to estimate a canonical avatar from a single image. It computes the Radial Basis Function (RBF) distance between query points to body landmarks as geometric features. ARCH++~\cite{he2021arch++} employs Pointnet++~\cite{qi2017pointnet++} as a geometry encoder to extract human body prior information. However, the geometric features leveraged from a naked body make the recovered surface lose details. The human body prior provides pose information which is helpful for reconstructing 3D humans in the original posed space (e.g. ICON), but has a minor effect for canonical shape recovery. The geometry cues such as normal are supposed to be more important for clothed avatar reconstruction. Different from ARCH and ARCH++, We utilize the canonical normal transformed from the original space to help preserve high-frequency details.  
 
\section{Method}
\label{sec:method} 
Figure~\ref{fig:overview} shows the framework of the clothed avatar reconstruction (CAR) method. Given an image, the front and back normal images and a SMPL body are simultaneously obtained by the body-guided normal prediction model described in~\cite{xiu2022icon}. In the first stage, the canonical implicit model takes the predicted normal image as input and recover a coarse human body in canonical space. In the second stage, the coarse mesh surface is implicitly refined by a SDF Network. 
 
\subsection{Canonical Implicit Model}\label{sec:cim} 
The canonical implicit model aims to reconstruct the general shape of a subject in the canonical space. Previous methods~\cite{huang2020arch, he2021arch++} estimate an occupancy field by learning a classification task that whether a 3D point is inside or outside a target human body. This scheme is unfriendly for animatable avatar reconstruction tasks where the mapping between posed space and canonical space is required. While in practice, imperfect mapping is unavoidable and there is always a gap between estimated poses and ground truth poses. As a result, the occupancy based methods tend to classify the points inside a human body as background when they are erroneously projected to the background area in a 2D image. 

Unlike the occupancy based methods which are point level, our method adopts a signed distance function (SDF) to approximate the target human body in a surface level, which is more robust to local mapping noises. Unlike occupancy, SDF aims to find an optimal surface where the surface normal is closest to the target surface normal. Instead of using the classification loss, CAR constrains points' gradients and surface normal as in~\cite{gropp2020igr}. Table \ref{tab:methods} lists a comparison of the implicit functions, which mainly use image features and geometric features to estimate a points' signed distance or occupancy value.

There are three kinds of features used in our implicit function $\mathcal{F}$ to predict a point's signed distance to a target surface, which are pixel-aligned image feature $\Phi^\dagger$, canonical normal \textbf{$\mathbf{n_c}$} and the canonical location $\mathbf{x_c} \in \mathbb{R}^3$. The zero-level surface is formulated as:  
\begin{align} 
\mathcal{S}_{\eta} = \{\mathbf{x_c}   \in \mathbb{R}^3 | \mathcal{F}(\Phi^\dagger, \mathbf{n_c}, \mathbf{x_c}; \eta)=0\},
\end{align}
where $\eta$ is the network parameters.  

\noindent \textbf{Linear Blend Skinning.} LBS \cite{SMPL:2015} is widely used to control the large-scale movements of a human body, by transforming the skin according to the motion of the skeleton. Let $\mathbf{X}=\{ \mathbf{x}_{\mathbf{c}}^{i}\in  \mathbb{R}^{3}\}_{i=1}^{N_\mathbf{v}}$ be the body vertices in the canonical space and $W=\{ w^{i} \in  \mathbb{R}^{N_J}\}_{i=1}^{N_\mathbf{v}}$ be the vertex-to-bone skinning weights, where $N_\mathbf{v}$ and $N_J$ are the number of vertices and joints respectively. For simplicity, we omit index $i$ for $\mathbf{x}_{\mathbf{c}}^{i}$ and $w^{i}$. Given pose parameters $\theta \in \mathbb{R}^{N_J\times 3}$ and joints $J$, the LBS function $\mathcal{W}$ maps a canonical vertex $\mathbf{x_c}$ with its skinning weight $w \in \mathbb{R}^{N_J}$ to the target posed space as follow:  
\begin{align}\label{eq:lbs}
    \mathbf{x_p}= \mathcal{W}(\mathbf{x_c},w,\theta,J) = \sum_{j=1}^{N_J} w_{j}\mathbf{B}_j(\theta, J) \mathbf{x_c},
\end{align} 
where $\mathbf{B}_j(\theta, J)$ is the bone transformation applied on a body part $j$. We define $\mathcal{W}^{-1}$ as the inverse LBS function mapping vertices from original space to canonical space.

\noindent \textbf{Pixel-Aligned Image Feature.} We use Stacked Hourglass (SHG) as the normal image encoder which is the same as in~\cite{saito2019pIFu, saito2020pifuhd, huang2020arch, he2021arch++}. Given an RGB image $\mathcal {I}\in \mathbb{R}^{H \times W \times 3} $, we predict the normal image $\mathcal {N}$ using the normal prediction model in \cite{xiu2022icon}. Then, the normal image encoder takes $\mathcal {N}$ as input and outputs a feature map $G(\mathcal {N}) \in \mathbb{R}^{H'\times W' \times C' }$. By projecting a point in posed space onto a normal image plane, the pixel-aligned image features can be obtained using the bilinear interpolation as follow: 
\begin{align} 
\Phi^\dagger=\mathcal{B} (G(\mathcal {N}),\pi(\mathbf{x_p}) ),
\end{align}
where $\mathbf{x_p}$ is the deformed point in the posed space obtained by Eqn.~\ref{eq:lbs}, $\pi(\cdot)$ indicates the weak orthogonal camera projection, and $\mathcal{B}$ denotes the bilinear interpolation operation.

\begin{table}[!t]
      \caption{
      Comparison of implicit functions of different human body reconstruction methods. $\Phi^\dagger$ denotes the pixel-aligned image feature, $\mathcal{N}^\dagger$ denotes the pixel-aligned normal predicted from an RGB image, $\Psi$ denotes the geometric feature leveraging the human body prior $\mathcal{M}$, $\mathbf{x}=(x,y,z)\in \mathbb{R}^3$ is a 3D point, $z$ is the depth of $\mathbf{x}$ and  $\mathbf{n}$ is the normal of $\mathbf{x}$ in the canonical space.
      } 
\renewcommand\arraystretch{0.6}   
      \resizebox{0.48\textwidth}{!} { 
      \begin{tabular}{l l }  
      \toprule[1.pt] 
     \textbf{\centering Method} & \textbf{Implicit Function}	\cr
    \midrule[1.pt]  
       PIFu~\cite{saito2019pIFu}, PIFuHD~\cite{saito2020pifuhd} & $\mathcal{F}(\Phi^\dagger, z)$  	\cr  
      PHORHUM~\cite{alldieck2022phorhum}   & $\mathcal{F}(\Phi^\dagger, \mathbf{x})$ 		\cr 
      	PaMIR~\cite{zheng2021pamir}, GeoPIFu\cite{he2020geopifu},ARCH~\cite{huang2020arch}  &$\mathcal{F}(\Phi^\dagger, \Psi(\mathcal{M}))$  
     	\cr 
        ICON~\cite{xiu2022icon}   &$\mathcal{F}(\mathcal{N}^\dagger, \Psi(\mathcal{M}))$  
     	\cr 
     CAR (ours)   &  $\mathcal{F}(  \Phi^\dagger, \mathbf{n}, \mathbf{x}) $\cr
    \bottomrule[1.2pt] 
      \end{tabular}
      } 
      \label{tab:methods}
\end{table}

\noindent \textbf{Geometric Feature.} CAR leverages canonical normal as the geometric feature. Inspired by \cite{seyb2019non}, the canonical normal $\mathbf{n_c}$ can be obtained by an inverse transform of the predicted normal $\mathbf{n_p}$ in original space, using the Jacobian matrix of a forward transform of $\mathbf{x_c}$.  
\begin{align} 
\mathbf{n_c}  = unit(\nabla_{\mathbf{x_c}}  \mathcal{W}^{-1} \cdot \pi^{-1} \cdot  \mathbf{n_p}),
\end{align} 
where $\pi^{-1}$ means the inverse camera projection matrix, $\mathbf{n_p}=\mathcal{B}(\mathcal{N},\pi(\mathbf{x_p}))$ is the pixel-aligned normal indexed from predicted normal image $\mathcal{N}$, and $unit(\cdot )$ means normalizing the input vector. Ablation study on different types of geometric features shows that the normal feature is better than other methods. 
  
\noindent \textbf{Point Position.} Specifically, we use the basic position encoding \cite{mildenhall2021nerf} of the canonical point as an additional feature. This term is maintained for the sake of computing gradients and surface normal $\triangledown_\mathbf{x} \mathcal{F}$.   
 
\noindent \textbf{Training Loss.} Following \cite{gropp2020igr}, our training loss contains three terms: surface reconstruction loss $\mathcal{L}_\mathrm{I}$, geometric regularization loss $\mathcal{L}_\mathrm{eik}$, and off-surface regularization $\mathcal{L}_{\mathrm{o}}$. 
\begin{align} 
\mathcal{L} = \lambda_\mathrm{I} \mathcal{L}_{\mathrm{I}} + \lambda_\mathrm{eik} \mathcal{L}_{ \mathrm{eik} } + \lambda_\mathrm{o} \mathcal{L}_{ \mathrm{o}},
\label{eq:all_loss}
 \end{align} 
Empirically, we set $\lambda_\mathrm{I}=1,\lambda_\mathrm{eik}=0.1,\lambda_\mathrm{o}=0.1$.

\noindent \textbf{Reconstruction Loss.} $\mathcal{L}_\mathrm{I}$ is a reconstruction loss, which ensures the Signed Distance Function vanish on surface points and its normal is consistent with the ground truth surface normal.   
\begin{align} 
\mathcal{L}_{\mathrm{I}}  = \frac{1}{|\Omega_\mathrm{I} |} \sum_{\mathbf{x}\in\Omega_\mathrm{I} }^{} (|\mathcal{F}_\mathbf{x}| +   ||\mathbf{n}_\mathbf{x}-\hat{\mathbf{n}}_\mathbf{x}  ||),
\label{eq:rec_loss}
\end{align} 
where $\mathbf{n}_\mathbf{x}=\triangledown_\mathbf{x} \mathcal{F}_\mathbf{x}$ is the differential normal at $\mathbf{x}$, $\hat{\mathbf{n}}_\mathbf{x}$ is the target normal and $\Omega_\mathrm{I}$ is a set of points which are randomly sampled from surface points.  

\noindent \textbf{Eikonal Loss.} The formulated Eikonal loss \cite{atzmon2020sal} is a regular loss commonly used to constrain $\mathcal{G}$ and $\mathcal{F}$ to be a SDF, by enforcing the implicit function to have a unit gradient:
\begin{align} 
\mathcal{L}_{ \mathrm{eik} } = \frac{1}{|\Omega_\mathrm{D}|}  \sum_{\mathbf{x}\in\Omega_\mathrm{D} }  (||\mathbf{n}_\mathbf{x}||-1), 
\label{eq:eik_loss}
\end{align} 
where $\Omega_\mathrm{D}$ is a point set sampled from a uniform distribution within the bounding box. 
 
\noindent \textbf{Off-surface Regularization.} $\mathcal{L}_{\mathrm{o}}$ enforces the sign distance of points far from the surface as large as possible as in \cite{sitzmann2020implicit}.
\begin{align} 
\mathcal{L}_{\mathrm{o}} = \frac{1}{|\Omega_\mathrm{D}|} \sum_{\mathbf{x}\in\Omega_\mathrm{D} }  \exp(-\alpha \cdot |\mathcal{F}_{\mathbf{x}}|),
\label{eq:reg_loss}
\end{align}  
where $\alpha \gg 1$ is the sharpness of decision boundary.

\subsection{Normal Refinement Process}\label{sec:refinement}  
The canonical implicit model tends to ignore surface details and focuses on the general shape of the captured subject due to the the bad mapping between canonical space and posed space. To solve this problem, we refine the predicted canonical surface and enforce it to be consistent with the input image. Like~\cite{Zheng2019DeepHuman, xiu2022icon}, we adopt the normal image in the posed space to refine the reconstructed results. Different from previous methods, we use an implicit function, i.e., a Signed Distance Function instead of learning the per-vertex displacement which always produce artifacts because the topology of a mesh is fixed. The normal refinement process mainly consists of two parts: a surface reconstruction network $\mathcal{G_\varphi}$ to generate a high-quality human body by a SDF, and a meta hyper-network $\mathcal{H_\phi}$  generating the initial parameters of the reconstruction network for fast optimization.  

\subsection{Hyper Network Training} 

Optimizing the reconstruction network $\mathcal{G_\varphi}$ from scratch is inefficient and not necessary, since a coarse mesh is already obtained in the first stage and it can be used as an initialization. Then the problem becomes how to estimate the parameters $\varphi_0$ that  $\mathcal{G}_{\varphi_0}$ can approximate a known mesh. Our solution is to leverage a hyper network $\mathcal{H_\phi}$ takes a mesh as input and output a set of parameters. We condition the hyper network on a SMPL body mesh instead of the reconstruction in the first stage. The are two main reasons. The first is that SMPL body mesh is naked without clothing variation, and the topology is simple and easy for learning. The second reason is that the real 3D human data is limited, while a large scale SMPL data with various poses and shapes can be synthesized to train the hyper network, thus improving the networks' generalization to unseen data. Note the hyper-network is only trained once in our method. 
    
Given a SMPL body $\mathcal{M}(\theta, \beta)$, the hyper network $\mathcal{H}_\phi$ generates a set of parameters $\varphi_0= \mathcal{H}(\mathcal{M};\phi )$, which are used to parameterize the SDF reconstruction network $\mathcal{G}$ to reproduce $\mathcal{M}$. The zero-level surface can be represented as follow:
\begin{align}
\mathcal{M}^* &= \{\mathbf{x}   \in \mathbb{R}^3 | \mathcal{G}(\mathbf{x}  ; \mathcal{H}(\mathcal{M};\phi ))=0\}
\end{align}
where $\phi$ are learnable parameters of the hyper network. During training, $\phi$ is updated by enforcing $\mathcal{M}^*$ closed to $\mathcal{M}$. The training loss is the same as equation (\ref{eq:all_loss}). Once the hyper network is trained, the output parameters are used to initialize the SDF reconstruction network as $\mathcal{G}_{\varphi_0}$.


\subsection{SDF Network Optimization} 
After obtaining the canonical mesh by the canonical implicit model described in section \ref{sec:cim}, it is warped to the posed space and refined to improve the quality of surface geometry. The zero isosurface in the posed space is formulated as:  
\begin{align} 
\mathcal{S_\varphi} = \{\mathbf{x}   \in \mathbb{R}^3 | \mathcal{G}(\mathbf{x}  ; \varphi)=0\},
\label{eq:surface optimization}
\end{align} 
where $\varphi$ is trainable. Starting from $\varphi_0$, $\varphi$ is updated by optimizing the surface normal supervised by the predicted normal image $\mathcal{N}$. The optimization loss is similar to equation (\ref{eq:all_loss}). For the second term in equation (\ref{eq:rec_loss}), the target normal $\hat{\mathbf{n}}_\mathbf{x}=\mathcal{B}(\mathcal{N},\pi(\mathbf{x_p}))$ can be obtained by projecting a point to either front or back normal image. In practice, we use both front and back normal images to optimize the surface normal. 

 
\begin{table*}[!t]
      \caption{Quantitative comparisons 
      of different methods in both canonical and posed space on MVP-Human (MVP) and RenderPeople (RP). 
      }
      \vspace{-0.5em}
\renewcommand\arraystretch{1.1} 
    \centering 
      \resizebox{\textwidth}{!} {
      \begin{tabular}{r||ccc||ccc||ccc||ccc} 
    \thickhline
      \multirow{2}{*}{\textbf{Methods}}  & \multicolumn{3}{c||}{\textbf{MVP-Canonical}} &
      \multicolumn{3}{c||}{\textbf{MVP-Posed}}  &
      \multicolumn{3}{c||}{\textbf{RP-Canonical}} &
      \multicolumn{3}{c}{\textbf{RP-Posed}}\\ 
    \Cline{2-13}
    & Chamfer$\downarrow$   & P2S$\downarrow$ & Normal$\downarrow$   & Chamfer$\downarrow$   & P2S$\downarrow$ & Normal$\downarrow$ &  Chamfer$\downarrow$   & P2S$\downarrow$ & Normal$\downarrow$ &  Chamfer$\downarrow$   & P2S$\downarrow$ & Normal$\downarrow$  \\ 
   \thickhline 
     PIFu~\cite{saito2019pIFu}&-&-&-& 4.9638 & 6.1931 & 0.8013 & - & - &-& 4.8884& 5.1182  &0.7089\\  
      PIFuHD~\cite{saito2020pifuhd}&- & - & - &3.9068&4.3833&0.8247  &- & - &- &  5.2701 & 5.3971 & 0.7375  \\
      ICON~\cite{xiu2022icon}  &   - & - & -  & 3.9583 & 4.3886   &0.1957 &  - & - & - &   4.9126 & 5.1269  & 0.7610  \\  
        ARCH~\cite{huang2020arch} &1.5894 &1.8044&\textbf{0.0942} & 3.8274&4.3614&0.1819  &  2.3916  &   2.1424  & 0.1178&  2.3225   &2.0506 &    0.1543   \\  
      ARCH++~\cite{he2021arch++} & 2.3906&2.0035  &0.1849 & 4.0438&3.9825&0.2523 &    1.9046 & 1.8306 & 0.0971 & 1.8805 &  1.7720 &  0.1065  \\   
    CAR (ours)  &\textbf{1.0572} & \textbf{1.0811} & 0.1287 & \textbf{1.0771}& \textbf{1.0654} &\textbf{0.0902 }&\textbf{1.5401} & \textbf{1.4963} & \textbf{0.0821} &\textbf{1.5142 }&\textbf{1.4147}& \textbf{0.0871}  \\  
     \thickhline   
      \end{tabular}
      }
    \label{tab:compare}
\end{table*}  

\begin{figure*}[!t]
\centering
\includegraphics[width=\linewidth]{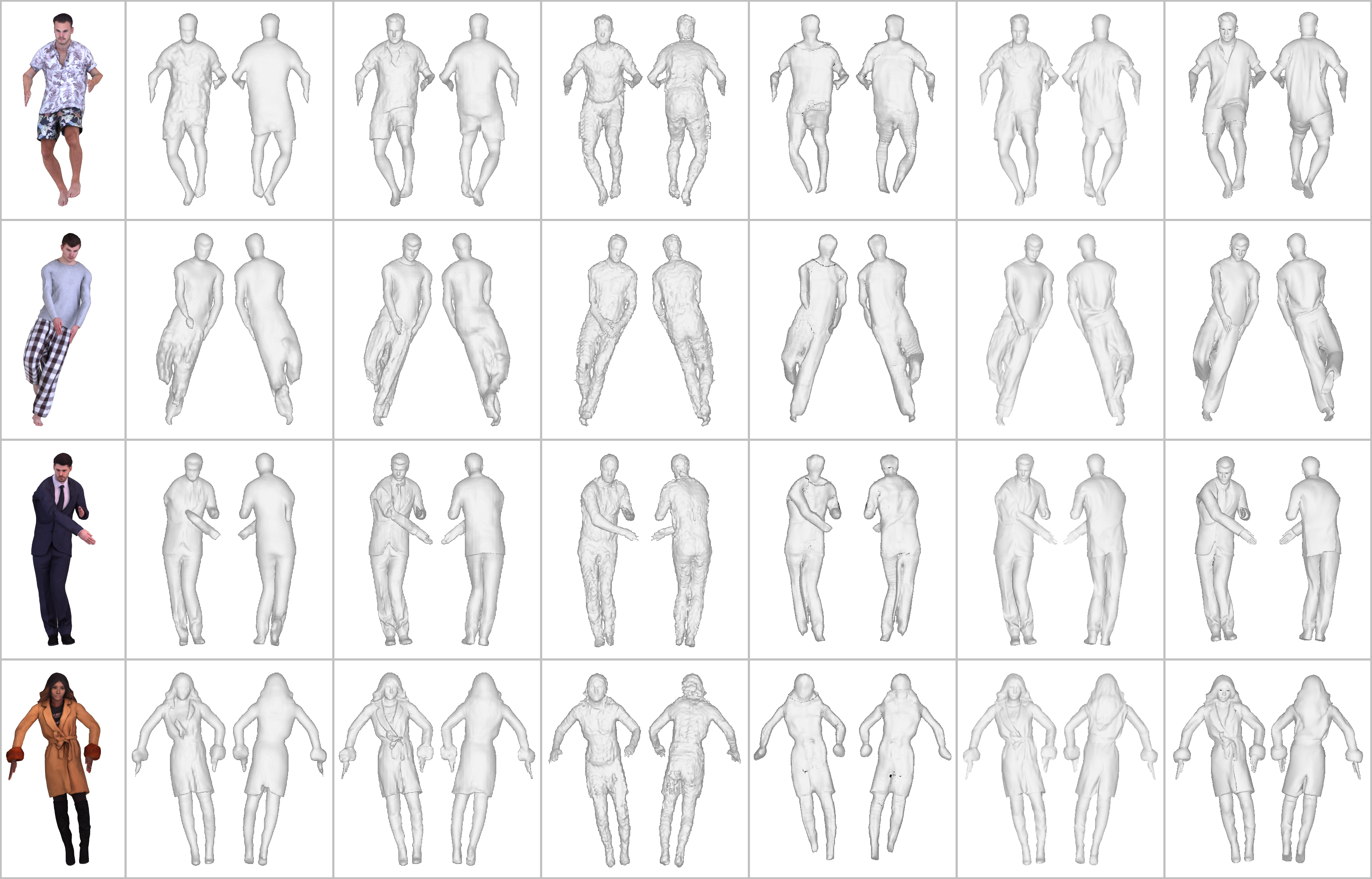} 
\rightline{\small Input \quad\quad\quad\quad  PIFu   \quad\quad\quad\quad\quad\,  PIFuHD   \quad\quad\quad\quad\quad\,  ICON 
\quad\quad\quad\quad\quad  ARCH  
\quad\quad\quad\quad\quad  CAR(Ours) 
\quad\quad\quad\quad\quad  GT \quad\quad\quad }
\vspace{-1.5 em}
\caption{Qualitative comparisons against the state-of-the-art methods on RenderPeople testing set. 
}
\label{fig:compare}
\vspace{-1.0 em}
\end{figure*}

\section{Experiments} 
In this section, we evaluate CAR with state-of-the-art methods on MVP-Human~\cite{zhu2022mvp} and RenderPeople~\cite{RenderPeople} datasets. The ablation study and discussion are also conducted to show the effectiveness of the proposed method.
  
\noindent \textbf{Dataset Description.} The training set contains 100 subjects from MVP-Human dataset and 50 subjects from RenderPeople dataset. The testing set includes 50 scans from MVP-Human, 11 scans from RenderPeople, and 2D real images from the internet. There is no intersection of training and testing sets. In the training phase, we fit a rigged 3D body template in the canonical pose with corresponding skinning weights to the scan mesh for each subject. We generate a motion sequence for each subject by warping the canonical mesh with the poses provided in AIST++ dataset \cite{aist-dance-db}. All generated meshes are rendered by rotating a camera around the vertical axis with intervals of 3 degrees.   

\noindent \textbf{Implementation Details.} We use stacked hourglass network \cite {newell2016stacked} as the normal image encoder which has the same architecture with \cite{he2020geopifu, saito2020pifuhd, huang2020arch}. The MLP of $\mathcal{F}$ has the number of neural layers (262, 512, 512, 512, 512, 512, 512, 1) with a skip connection at the fourth layer. We use the geometric initialization proposed in \cite{atzmon2020sal} for $\mathcal{F}$. The SMPL encoder has the number of neural layers (6, 256, 256, 256, 256, 256) with skip connections at the second, the third and the fourth layer, while the decoder contains 5 blocks, each has 3 hidden layers with 256 neural layers and an output layer whose number of neural is the same as the parameters' number of the corresponding layer of SDF network. The SDF network has the number of neural layers (3, 1024, 512, 256, 128, 1) and the parameters of each layer are initialized by the output of each decoder block. The canonical implicit model is trained with a batch size of 4 and a random window crop of 512 × 512 sizes. We use Adam optimizer and learning rate 1e-3 with decay by a factor of 0.1 every 3 epochs. In each iteration, we sample 8192 points on the surface, 8192 points around the surface with a normal distribution sigma of 0.1, and 2048 points uniformly sampled in a bounding box. We train the hyper-network using SMPL data generate by shape and pose parameters randomly sampled from SMPL data distribution. The normal prediction network and SMPL estimation are followed from \cite{xiu2022icon}.
 
\subsection{Quantitative Evaluation}
We compare our method with two kinds of methods: clothed human reconstruction algorithms include PIFu~\cite{saito2019pIFu}, PIFuHD~\cite{saito2020pifuhd} and ICON~\cite{xiu2022icon}, and avatar reconstruction approaches ARCH~\cite{huang2020arch} and ARCH++~\cite{he2021arch++}. Table \ref{tab:compare} illustrates the quantitative results in both canonical and posed space on MVP-Human and RenderPeople datasets. For PIFu, PIFuHD and ICON, we directly use the model published by original works. For ARCH and ARCH++, we train the model by ourselves in our training set and report the testing performance. From the results, although PIFu/PIFuHD usually shows good performance in visual results, they do not look so good in quantitative evaluation.
That's because they do not utilize the human body prior and are not so robust to pose variations.
ICON leverages SMPL to improve pose robustness. However, it relies too heavily on the naked SMPL body and is not good enough to handle loose clothes such as coats (4th column, 4th row in Figure~\ref{fig:compare}).
ARCH, ARCH++, and the proposed CAR introduce the process of general shape reconstruction in the canonical space so that the pose robustness is greatly improved. Our method CAR further pays more attention to the geometry detail recovery on the surface, and it achieves the best accuracy over all datasets in both canonical and posed space, validating its effectiveness for the clothed avatar reconstruction. 

\begin{figure*}[!t]
\centering
\scriptsize
\includegraphics[width=\linewidth]{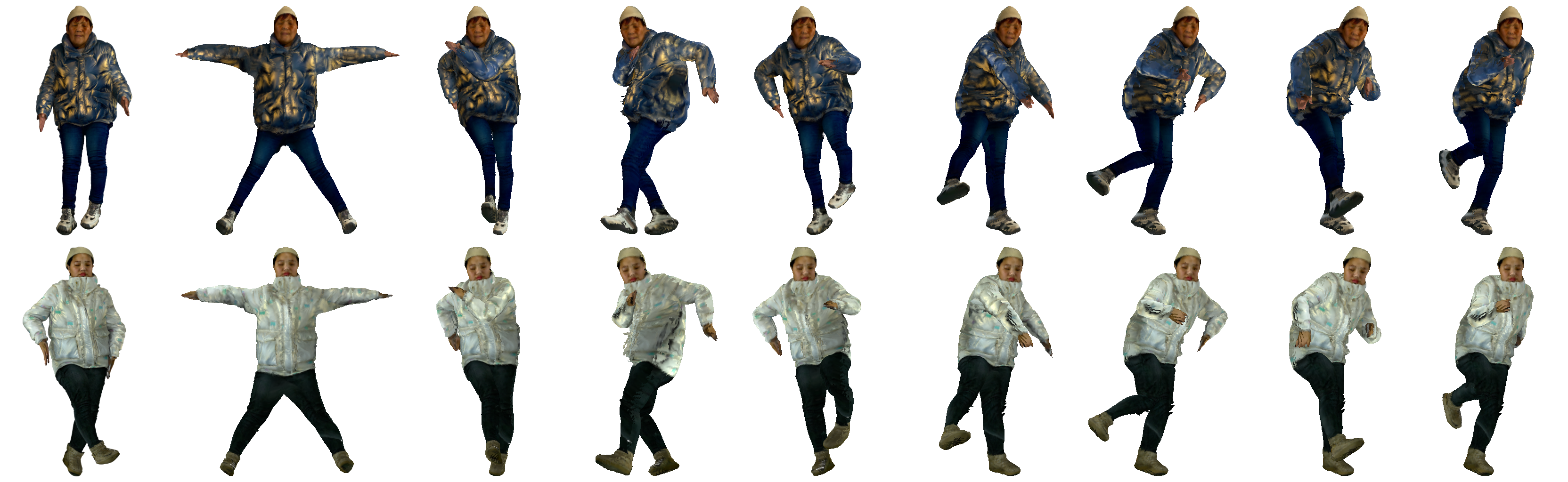} 
\leftline { \small  \, Input Image \quad  Canonical Result \quad\quad\quad\quad\quad\quad\quad\quad\quad\quad\quad\quad\quad\quad\quad\quad   Animated Avatar    }  
\vspace{-1.5 em}
\caption{Images to animated avatars. 
}
\vspace{-0.5 em} 
\label{fig:avatar}
\end{figure*} 

\begin{figure*}[!t]
\centering
\includegraphics[width=\linewidth]{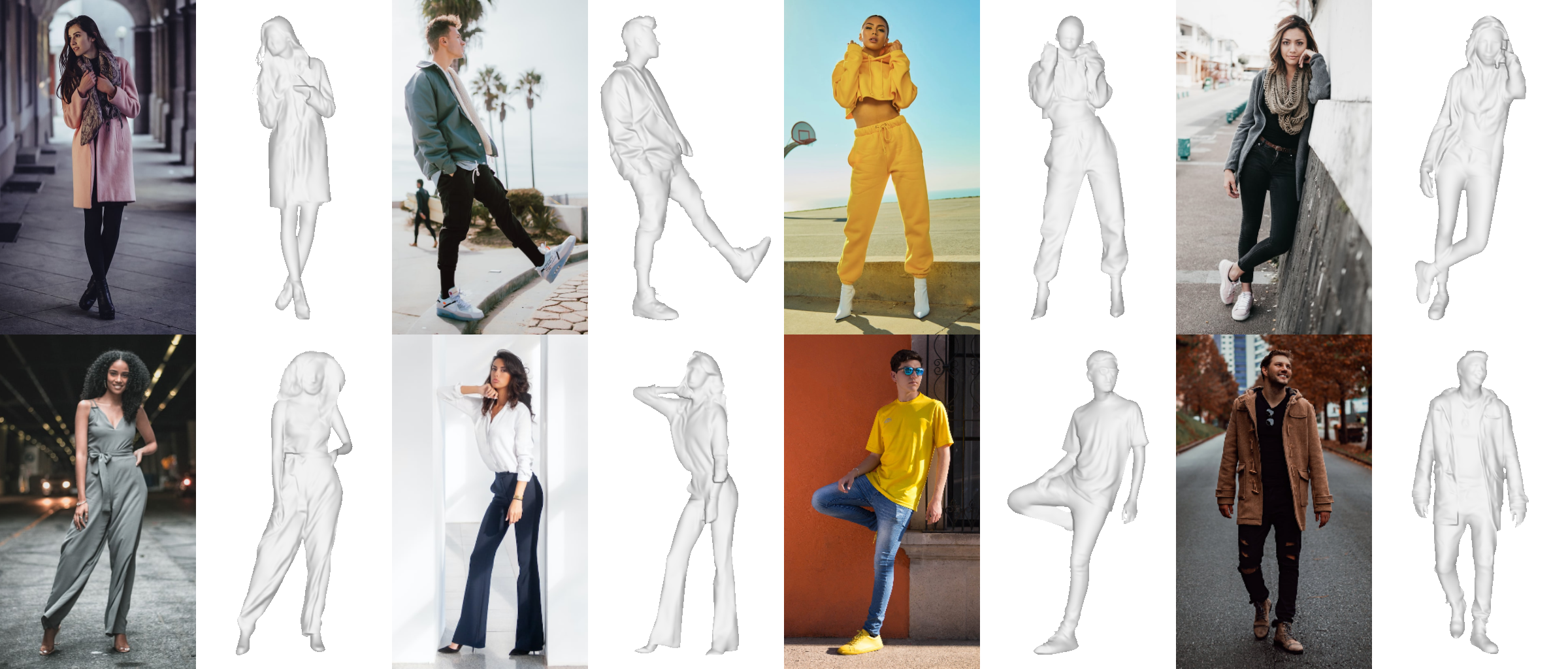}  
\vspace{-1.5 em}
\caption{Qualitative results on real images from the internet. These results demonstrate that our model trained by synthetically generated
data can successfully reconstruct high-fidelity 3D from humans in real world data.}
\vspace*{-0.5 em} 
\label{fig:real_image}
\end{figure*}
   
\begin{table}[!t]
      \caption{Ablation study of different modules (RenderPeople).}
       \vspace{-0.5em}
\renewcommand\arraystretch{1.2} 
    \centering 
      \resizebox{0.48\textwidth}{!} {
      \begin{tabular}{r||ccc||ccc} 
    \thickhline
      \multirow{2}{*}{\textbf{Methods}}  &  
      \multicolumn{3}{c||}{\textbf{Canonical Space}} &
      \multicolumn{3}{c}{\textbf{Posed Space}}\\ 
    \Cline{2-7}
     & Chamfer$\downarrow$   & P2S$\downarrow$ & Normal$\downarrow$ &  Chamfer$\downarrow$   & P2S$\downarrow$ & Normal$\downarrow$  \\ 
   \thickhline  
      CAR,baseline   &1.5760 &1.5766 &\textbf{0.0774} & 1.5484 &1.4926 &0.0983  \\  
      +GeoFeat  &   \textbf{1.5329} &1.5069 &  0.0927 &\textbf{1.5043}   & 1.4239 &0.1120 \\  
      +Refine  &1.5401 & \textbf{1.4963} & 0.0821 &1.5142&\textbf{1.4147}& \textbf{0.0871}  \\  
       
      \thickhline
      \end{tabular}
      }
      \vspace{-1.0 em}
    \label{tab:ablation} 
\end{table}

\subsection{Qualitative Evaluation}
\Cref{fig:compare} shows the qualitative results in RenderPeople. PIFu fails to reconstruct whole limbs since it does not use human body priors. PIFuHD captures better details than PIFu, but the backside of reconstructions is overly smooth due to the lacking of end-to-end geometry encoder. ICON suffers from surface noise since the image global encoder is removed thus extracted features are purely local. It is worth noting that PIFu, PIFuHD and ICON do not support animation. ARCH can generate animatable avatars, but the recovered surface is overly smooth or with artifacts. Our method successfully produces realistic 3D humans which can be animated, see~\cref{fig:avatar} for some examples. ~\Cref{fig:real_image} shows more results on in-the-wild images, which demonstrates that our method can reconstruct high-fidelity 3D humans, regardless of poses or clothing.

\begin{figure*}[!t]
\centering
\includegraphics[width=\linewidth]{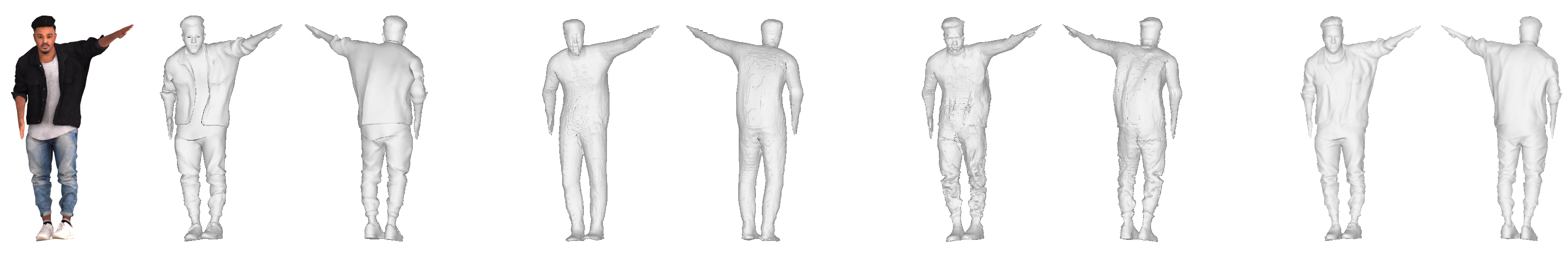} 
\rightline {  \small
Input\quad\quad\quad\quad\quad\quad  
GT \quad\quad\quad\quad\quad\quad\quad\quad\quad
Ours w/o GeoFeat   \quad\quad\quad\quad\quad  
Ours w/o refinement  \quad\quad\quad\quad\quad\quad\quad\quad 
Ours  \quad\quad\quad\quad } 
\vspace{-1.5 em}
\caption{Ablation analysis on different modules, corresponding to \cref{tab:ablation}.}
\vspace*{-1.0 em} 
\label{fig:ablation}
\end{figure*}

\begin{table*}[t]
      \caption{ 
     Reconstruction errors on different types of loss and geometric feature variants of our method on MVP-Human.
      } 
       \vspace{-1.0em}
\renewcommand\arraystretch{1.1} 
    \centering 
      \resizebox{\textwidth}{!} {
      \begin{tabular}{c||l||ccc||ccc||ccc}
      \thickhline
       \multirow{2}{*}[-0.5ex]{}  &
      \multirow{2}{*}[-0.5ex]{\textbf{Methods}}  & 
      \multicolumn{3}{c||}{\textbf{Canonical Space}} &
      \multicolumn{3}{c||}{\textbf{Posed Space}} &
      \multicolumn{3}{c}{\textbf{Mean}}\\
    \Cline{3-11}
 &  &   Chamfer$\downarrow$ & P2S$\downarrow$ &  Normal$\downarrow$   &Chamfer$\downarrow$ & P2S$\downarrow$ &  Normal$\downarrow$ & Chamfer$\downarrow$ & P2S$\downarrow$ &  Normal$\downarrow$ \\
      \Hline 
    \multirow{2}{*}{\textbf{Loss}}  & occupancy     & 1.3867   & 1.4974   & 0.1119&   3.5136  & 3.9336   &0.1945  &  2.4501    &  2.7155  & 0.1532 \\ 
     
    &sdf (ours) &   1.3758 &1.4081   & 0.0896& 3.7589  & 3.8199   &0.2006    & 2.5673 & 2.6140   & 0.1451       \\  
     \Hline 
    \multirow{3}{*}{\textbf{Geometric Feature}}  &   P2J distance &   1.5227  & 1.3106 & 0.1288&  3.8324   &  3.7256  & 0.2310     & 2.6775&2.5181     &0.1799  \\ 
       &  signed distance      & 2.0142 &  1.3137 &0.1537&  4.2412 & 3.7999  & 0.2365 & 6.5639 &  2.3932  & 0.3429 \\  
         & surface normal (ours)   &1.2463 & 1.3020  &0.1053   &     3.6058&  3.7560& 0.2069     & 2.4411   & 2.5237  &0.1570   \\
     \Hline   
      \end{tabular}
      }
      \vspace{-1.0 em}
    \label{tab:analysis} 
\end{table*} 

\subsection{Analysis and Discussion}
\label{sec:analysis}
To evaluate the influences of different factors, we conduct three experiments including: 1) the ablation study on our method of different factors; 2) the comparison of SDF and occupancy losses; 3) different choices of geometric features.  

\smallskip
\noindent \textbf{Ablation Study.} Table~\ref{tab:ablation} and Figure~\ref{fig:ablation} demonstrate the effectiveness of different parts of our method. The first row of Table \ref{tab:ablation} is our method without the geometric feature and normal refinement. We denote it $\mathcal{F}(\Phi^\dagger,\mathbf{x_c})$ as the baseline method. 
The second row is our method described in section \ref{sec:cim} without the normal refinement process. The third is our proposed method. We can see that the geometry feature improves the accuracy of reconstructed results. The normal refinement process further reduces the reconstruction errors and their qualitative results are shown in Figure~\ref{fig:ablation}.

\noindent \textbf{Occupancy vs SDF.} Previous methods \cite{saito2019pIFu, saito2020pifuhd, xiu2022icon, huang2020arch, he2021arch++} tries to estimate an occupancy value for a query point that "1" means inside the human body while "0" means outside. To this end, a regression loss (e.g., L2 loss) or classification loss (e.g., binary cross entropy loss) is always used to enforce estimations to be close to the real occupancy field. Our method, instead, predicts a SDF field and trains our network using the loss described in section \ref{sec:cim}. In this part, we compare the results of these two kinds of losses. We train two models using L2 occupancy loss and SDF loss respectively. For a fair comparison, all configurations are the same except the training loss. The first two rows of \cref{tab:analysis} show that the SDF loss performs better than occupancy loss.

\noindent \textbf{Geometric Feature Evaluation.} We analyse three different geometric features in~\cref{tab:analysis}: 1) spatial feature P2J distance proposed in ARCH~\cite{huang2020arch} (i.e., distance from a point to SMPL joints) 2) signed distance proposed in ICON~\cite{xiu2022icon} (i.e., distance from a point to the nearest point on SMPL surface) 3) ours with canonical normal. The bottom three rows in \cref{tab:analysis} show that the canonical normal performs the best, which achieves the lowest errors. 
 
\subsection{Inference Speed}  
The proposed CAR consists of two stages. The first stage adopts a learning-based way so that the shape recovery in canonical space is efficient. The second stage is an optimized way. Fortunately, with the initialization of the hyper network, the normal refinement process converges faster. It takes about 1500 iterations to output an optimal result (compared to 3000 iterations with random initialization). For a single subject, these two stages require an average time of about 5 minutes on a single TITAN X GPU, while optimization-based methods usually takes several hours to construct a subject.

\section{Conclusion}
\label{sec:conclusion} 
We present a method for clothed avatar reconstruction from a single image in free viewpoints and unconstrained poses. The person image is decomposed into a canonical mesh describing its general shape as well as a pose-dependent non-rigid deformation. By incorporating the normal information in both canonical mesh learning and the non-rigid deformation refinement process, we successfully reconstruct high-fidelity avatars which preserve surface details like cloth wrinkles. With a hyper network for parameter initialization, it further accelerates the convergent process and improves the optimization efficiency. Our method can be easily extended to multiple image settings and the avatar reconstruction results are expected to be improved. How to utilize consistent information across temporal images or monocular video of a dynamic human to reconstruct a complete avatar is one of our future works.  
\section*{Acknowledgement}
This work was supported in part by Chinese National Natural Science Foundation Projects \#62176256, \#62206280, \#62276254, the Tencent AI Lab Rhino-Bird Focused Research Program RBFR2022010, the OPPO Research Fund and the InnoHK program. Yuliang Xiu has received funding from the European Union’s Horizon $2020$ research and innovation programme under the Marie Skłodowska-Curie grant agreement No.$860768$ (\href{https://www.clipe-itn.eu}{CLIPE} project). Hongwei Yi is supported by the German Federal Ministry of Education and Research (BMBF): Tübingen AI Center, FKZ: 01IS18039B.

\clearpage
{\small
\balance
\bibliographystyle{config/ieee_fullname}
\bibliography{config/BIB}

\begin{thebibliography}{10}\itemsep=-1pt

\bibitem{RenderPeople}
Renderpeople.
\newblock \url{renderpeople.com}.

\bibitem{alldieck2018video}
Thiemo Alldieck, Marcus Magnor, Weipeng Xu, Christian Theobalt, and Gerard
  Pons-Moll.
\newblock Video based reconstruction of 3d people models.
\newblock In {\em {IEEE} Conference on Computer Vision and Pattern
  Recognition}.
\newblock {CVPR} Spotlight Paper.

\bibitem{alldieck2018detailed}
Thiemo Alldieck, Marcus Magnor, Weipeng Xu, Christian Theobalt, and Gerard
  Pons-Moll.
\newblock Detailed human avatars from monocular video.
\newblock In {\em International Conference on 3D Vision}, pages 98--109, Sep
  2018.

\bibitem{alldieck2022phorhum}
Thiemo Alldieck, Mihai Zanfir, and Cristian Sminchisescu.
\newblock Photorealistic monocular 3d reconstruction of humans wearing
  clothing.
\newblock In {\em Proceedings of the IEEE/CVF Conference on Computer Vision and
  Pattern Recognition (CVPR)}, 2022.

\bibitem{alldieck2022photorealistic}
Thiemo Alldieck, Mihai Zanfir, and Cristian Sminchisescu.
\newblock Photorealistic monocular 3d reconstruction of humans wearing
  clothing.
\newblock In {\em Proceedings of the IEEE/CVF Conference on Computer Vision and
  Pattern Recognition}, pages 1506--1515, 2022.

\bibitem{atzmon2020sal}
Matan Atzmon and Yaron Lipman.
\newblock Sal: Sign agnostic learning of shapes from raw data.
\newblock In {\em Proceedings of the IEEE/CVF Conference on Computer Vision and
  Pattern Recognition}, pages 2565--2574, 2020.

\bibitem{balan2007detailed}
Alexandru~O Balan, Leonid Sigal, Michael~J Black, James~E Davis, and Horst~W
  Haussecker.
\newblock Detailed human shape and pose from images.
\newblock In {\em 2007 IEEE Conference on Computer Vision and Pattern
  Recognition}, pages 1--8. IEEE, 2007.

\bibitem{bhatnagar2020ipnet}
Bharat~Lal Bhatnagar, Cristian Sminchisescu, Christian Theobalt, and Gerard
  Pons-Moll.
\newblock Combining implicit function learning and parametric models for 3d
  human reconstruction.
\newblock In {\em European Conference on Computer Vision ({ECCV})}. {Springer},
  August 2020.

\bibitem{bhatnagar2020loopreg}
Bharat~Lal Bhatnagar, Cristian Sminchisescu, Christian Theobalt, and Gerard
  Pons-Moll.
\newblock Loopreg: Self-supervised learning of implicit surface
  correspondences, pose and shape for 3d human mesh registration.
\newblock {\em Advances in Neural Information Processing Systems},
  33:12909--12922, 2020.

\bibitem{bhatnagar2019multi}
Bharat~Lal Bhatnagar, Garvita Tiwari, Christian Theobalt, and Gerard Pons-Moll.
\newblock Multi-garment net: Learning to dress 3d people from images.
\newblock In {\em Proceedings of the IEEE International Conference on Computer
  Vision (ICCV)}, pages 5420--5430, 2019.

\bibitem{chen2021snarf}
Xu Chen, Yufeng Zheng, Michael~J Black, Otmar Hilliges, and Andreas Geiger.
\newblock Snarf: Differentiable forward skinning for animating non-rigid neural
  implicit shapes.
\newblock In {\em Proceedings of the IEEE/CVF International Conference on
  Computer Vision}, pages 11594--11604, 2021.

\bibitem{deng2020nasa}
Boyang Deng, John~P Lewis, Timothy Jeruzalski, Gerard Pons-Moll, Geoffrey
  Hinton, Mohammad Norouzi, and Andrea Tagliasacchi.
\newblock Nasa neural articulated shape approximation.
\newblock In {\em European Conference on Computer Vision}, pages 612--628.
  Springer, 2020.

\bibitem{esteban2004silhouette}
Carlos~Hern{\'a}ndez Esteban and Francis Schmitt.
\newblock Silhouette and stereo fusion for 3d object modeling.
\newblock {\em Computer Vision and Image Understanding}, 96(3):367--392, 2004.

\bibitem{furukawa2006carved}
Yasutaka Furukawa and Jean Ponce.
\newblock Carved visual hulls for image-based modeling.
\newblock In {\em European Conference on Computer Vision}, pages 564--577.
  Springer, 2006.

\bibitem{gall2009motion}
Juergen Gall, Carsten Stoll, Edilson De~Aguiar, Christian Theobalt, Bodo
  Rosenhahn, and Hans-Peter Seidel.
\newblock Motion capture using joint skeleton tracking and surface estimation.
\newblock In {\em 2009 IEEE Conference on Computer Vision and Pattern
  Recognition}, pages 1746--1753. Ieee, 2009.

\bibitem{gropp2020igr}
Amos Gropp, Lior Yariv, Niv Haim, Matan Atzmon, and Yaron Lipman.
\newblock Implicit geometric regularization for learning shapes.
\newblock {\em arXiv preprint arXiv:2002.10099}, 2020.

\bibitem{habermann2019livecap}
Marc Habermann, Weipeng Xu, Michael Zollhoefer, Gerard Pons-Moll, and Christian
  Theobalt.
\newblock Livecap: Real-time human performance capture from monocular video.
\newblock {\em ACM Transactions On Graphics (TOG)}, 38(2):1--17, 2019.

\bibitem{habermann2020deepcap}
Marc Habermann, Weipeng Xu, Michael Zollhofer, Gerard Pons-Moll, and Christian
  Theobalt.
\newblock Deepcap: Monocular human performance capture using weak supervision.
\newblock In {\em Proceedings of the IEEE/CVF Conference on Computer Vision and
  Pattern Recognition}, pages 5052--5063, 2020.

\bibitem{he2020geopifu}
Tong He, John Collomosse, Hailin Jin, and Stefano Soatto.
\newblock Geo-pifu: Geometry and pixel aligned implicit functions for
  single-view human reconstruction.
\newblock In {\em Conference on Neural Information Processing Systems
  (NeurIPS)}, 2020.

\bibitem{he2021arch++}
Tong He, Yuanlu Xu, Shunsuke Saito, Stefano Soatto, and Tony Tung.
\newblock Arch++: Animation-ready clothed human reconstruction revisited.
\newblock In {\em Proceedings of the IEEE/CVF International Conference on
  Computer Vision}, pages 11046--11056, 2021.

\bibitem{hu3DBodyNet}
Pengpeng Hu, Edmond Ho, and Adrian Munteanu.
\newblock 3dbodynet: Fast reconstruction of 3d animatable human body shape from
  a single commodity depth camera.
\newblock {\em IEEE Transactions on Multimedia}, PP:1--1, 04 2021.

\bibitem{huang2020arch}
Zeng Huang, Yuanlu Xu, Christoph Lassner, Hao Li, and Tony Tung.
\newblock Arch: Animatable reconstruction of clothed humans.
\newblock In {\em Proceedings of the IEEE/CVF Conference on Computer Vision and
  Pattern Recognition}, pages 3093--3102, 2020.

\bibitem{jiang2022selfrecon}
Boyi Jiang, Yang Hong, Hujun Bao, and Juyong Zhang.
\newblock Selfrecon: Self reconstruction your digital avatar from monocular
  video.
\newblock In {\em Proceedings of the IEEE/CVF Conference on Computer Vision and
  Pattern Recognition}, pages 5605--5615, 2022.

\bibitem{jiang2020bcnet}
Boyi Jiang, Juyong Zhang, Yang Hong, Jinhao Luo, Ligang Liu, and Hujun Bao.
\newblock Bcnet: Learning body and cloth shape from a single image.
\newblock In {\em European Conference on Computer Vision}, pages 18--35.
  Springer, 2020.

\bibitem{lazova2019360}
Verica Lazova, Eldar Insafutdinov, and Gerard Pons-Moll.
\newblock 360-degree textures of people in clothing from a single image.
\newblock In {\em 2019 International Conference on 3D Vision (3DV)}, pages
  643--653. IEEE, 2019.

\bibitem{li2020monoportRTL}
Ruilong Li, Kyle Olszewski, Yuliang Xiu, Shunsuke Saito, Zeng Huang, and Hao
  Li.
\newblock Volumetric human teleportation.
\newblock In {\em ACM SIGGRAPH 2020 Real-Time Live!}, pages 1--1. 2020.

\bibitem{li2020monocular}
Ruilong Li, Yuliang Xiu, Shunsuke Saito, Zeng Huang, Kyle Olszewski, and Hao
  Li.
\newblock Monocular real-time volumetric performance capture.
\newblock In {\em European Conference on Computer Vision}, pages 49--67.
  Springer, 2020.

\bibitem{li2022avatarcap}
Zhe Li, Zerong Zheng, Hongwen Zhang, Chaonan Ji, and Yebin Liu.
\newblock Avatarcap: Animatable avatar conditioned monocular human volumetric
  capture.
\newblock In {\em ECCV}, 2022.

\bibitem{SMPL:2015}
Matthew Loper, Naureen Mahmood, Javier Romero, Gerard Pons-Moll, and Michael~J.
  Black.
\newblock {SMPL}: A skinned multi-person linear model.
\newblock {\em ACM Trans. Graphics (Proc. SIGGRAPH Asia)}, 34(6):248:1--248:16,
  Oct. 2015.

\bibitem{mildenhall2021nerf}
Ben Mildenhall, Pratul~P Srinivasan, Matthew Tancik, Jonathan~T Barron, Ravi
  Ramamoorthi, and Ren Ng.
\newblock Nerf: Representing scenes as neural radiance fields for view
  synthesis.
\newblock {\em Communications of the ACM}, 65(1):99--106, 2021.

\bibitem{newcombe2015dynamicfusion}
Richard~A Newcombe, Dieter Fox, and Steven~M Seitz.
\newblock Dynamicfusion: Reconstruction and tracking of non-rigid scenes in
  real-time.
\newblock In {\em Proceedings of the IEEE conference on computer vision and
  pattern recognition}, pages 343--352, 2015.

\bibitem{newell2016stacked}
Alejandro Newell, Kaiyu Yang, and Jia Deng.
\newblock Stacked hourglass networks for human pose estimation.
\newblock In {\em European conference on computer vision}, pages 483--499.
  Springer, 2016.

\bibitem{peng2021neural}
Sida Peng, Yuanqing Zhang, Yinghao Xu, Qianqian Wang, Qing Shuai, Hujun Bao,
  and Xiaowei Zhou.
\newblock Neural body: Implicit neural representations with structured latent
  codes for novel view synthesis of dynamic humans.
\newblock In {\em Proceedings of the IEEE/CVF Conference on Computer Vision and
  Pattern Recognition}, pages 9054--9063, 2021.

\bibitem{pons2017clothcap}
Gerard Pons-Moll, Sergi Pujades, Sonny Hu, and Michael~J Black.
\newblock Clothcap: Seamless 4d clothing capture and retargeting.
\newblock {\em ACM Transactions on Graphics (ToG)}, 36(4):1--15, 2017.

\bibitem{qi2017pointnet++}
Charles~Ruizhongtai Qi, Li Yi, Hao Su, and Leonidas~J Guibas.
\newblock Pointnet++: Deep hierarchical feature learning on point sets in a
  metric space.
\newblock {\em Advances in neural information processing systems}, 30, 2017.

\bibitem{saito2019pIFu}
Shunsuke Saito, Zeng Huang, Ryota Natsume, Shigeo Morishima, Angjoo Kanazawa,
  and Hao Li.
\newblock Pifu: Pixel-aligned implicit function for high-resolution clothed
  human digitization.
\newblock In {\em Proceedings of the IEEE/CVF International Conference on
  Computer Vision}, pages 2304--2314, 2019.

\bibitem{saito2020pifuhd}
Shunsuke Saito, Tomas Simon, Jason Saragih, and Hanbyul Joo.
\newblock Pifuhd: Multi-level pixel-aligned implicit function for
  high-resolution 3d human digitization.
\newblock In {\em CVPR}, 2020.

\bibitem{saito2021scanimate}
Shunsuke Saito, Jinlong Yang, Qianli Ma, and Michael~J Black.
\newblock Scanimate: Weakly supervised learning of skinned clothed avatar
  networks.
\newblock In {\em Proceedings of the IEEE/CVF Conference on Computer Vision and
  Pattern Recognition}, pages 2886--2897, 2021.

\bibitem{santesteban2019learning}
Igor Santesteban, Miguel~A Otaduy, and Dan Casas.
\newblock Learning-based animation of clothing for virtual try-on.
\newblock {\em Computer Graphics Forum (CGF)}, 38(2):355--366, 2019.

\bibitem{seyb2019non}
Dario Seyb, Alec Jacobson, Derek Nowrouzezahrai, and Wojciech Jarosz.
\newblock Non-linear sphere tracing for rendering deformed signed distance
  fields.
\newblock {\em ACM Transactions on Graphics}, 38(6), 2019.

\bibitem{sitzmann2020implicit}
Vincent Sitzmann, Julien Martel, Alexander Bergman, David Lindell, and Gordon
  Wetzstein.
\newblock Implicit neural representations with periodic activation functions.
\newblock {\em Advances in Neural Information Processing Systems},
  33:7462--7473, 2020.

\bibitem{slavcheva2017killingfusion}
Miroslava Slavcheva, Maximilian Baust, Daniel Cremers, and Slobodan Ilic.
\newblock Killingfusion: Non-rigid 3d reconstruction without correspondences.
\newblock In {\em Proceedings of the IEEE Conference on Computer Vision and
  Pattern Recognition}, pages 1386--1395, 2017.

\bibitem{starck2007surface}
Jonathan Starck and Adrian Hilton.
\newblock Surface capture for performance-based animation.
\newblock {\em IEEE computer graphics and applications}, 27(3):21--31, 2007.

\bibitem{tang2019neural}
Sicong Tang, Feitong Tan, Kelvin Cheng, Zhaoyang Li, Siyu Zhu, and Ping Tan.
\newblock A neural network for detailed human depth estimation from a single
  image.
\newblock In {\em Proceedings of the IEEE/CVF International Conference on
  Computer Vision}, pages 7750--7759, 2019.

\bibitem{aist-dance-db}
Shuhei Tsuchida, Satoru Fukayama, Masahiro Hamasaki, and Masataka Goto.
\newblock Aist dance video database: Multi-genre, multi-dancer, and
  multi-camera database for dance information processing.
\newblock In {\em Proceedings of the 20th International Society for Music
  Information Retrieval Conference, {ISMIR} 2019}, pages 501--510, Delft,
  Netherlands, Nov. 2019.

\bibitem{vlasic2008articulated}
Daniel Vlasic, Ilya Baran, Wojciech Matusik, and Jovan Popovi{\'c}.
\newblock Articulated mesh animation from multi-view silhouettes.
\newblock In {\em ACM SIGGRAPH 2008 papers}, pages 1--9. 2008.

\bibitem{wang2021metaavatar}
Shaofei Wang, Marko Mihajlovic, Qianli Ma, Andreas Geiger, and Siyu Tang.
\newblock Metaavatar: Learning animatable clothed human models from few depth
  images.
\newblock {\em Advances in Neural Information Processing Systems},
  34:2810--2822, 2021.

\bibitem{weng2019photo}
Chung-Yi Weng, Brian Curless, and Ira Kemelmacher-Shlizerman.
\newblock Photo wake-up: 3d character animation from a single photo.
\newblock In {\em Proceedings of the IEEE/CVF Conference on Computer Vision and
  Pattern Recognition}, pages 5908--5917, 2019.

\bibitem{xiang2020monoclothcap}
Donglai Xiang, Fabian Prada, Chenglei Wu, and Jessica Hodgins.
\newblock Monoclothcap: Towards temporally coherent clothing capture from
  monocular rgb video.
\newblock In {\em 2020 International Conference on 3D Vision (3DV)}, pages
  322--332. IEEE, 2020.

\bibitem{xiu2023econ}
Yuliang Xiu, Jinlong Yang, Xu Cao, Dimitrios Tzionas, and Michael~J. Black.
\newblock {ECON: Explicit Clothed humans Obtained from Normals}.
\newblock In {\em Proceedings of the IEEE/CVF Conference on Computer Vision and
  Pattern Recognition (CVPR)}, June 2023.

\bibitem{xiu2022icon}
Yuliang Xiu, Jinlong Yang, Dimitrios Tzionas, and Michael~J. Black.
\newblock {ICON}: {I}mplicit {C}lothed humans {O}btained from {N}ormals.
\newblock In {\em Proceedings of the IEEE/CVF Conference on Computer Vision and
  Pattern Recognition (CVPR)}, pages 13296--13306, June 2022.

\bibitem{xu2021h}
Hongyi Xu, Thiemo Alldieck, and Cristian Sminchisescu.
\newblock H-nerf: Neural radiance fields for rendering and temporal
  reconstruction of humans in motion.
\newblock {\em Advances in Neural Information Processing Systems},
  34:14955--14966, 2021.

\bibitem{xu2018monoperfcap}
Weipeng Xu, Avishek Chatterjee, Michael Zollh{\"o}fer, Helge Rhodin, Dushyant
  Mehta, Hans-Peter Seidel, and Christian Theobalt.
\newblock Monoperfcap: Human performance capture from monocular video.
\newblock {\em ACM Transactions on Graphics (ToG)}, 37(2):1--15, 2018.

\bibitem{yi2022mime}
Hongwei Yi, Chun-Hao~P. Huang, Shashank Tripathi, Lea Hering, Justus Thies, and
  Michael~J. Black.
\newblock {MIME}: Human-aware {3D} scene generation.
\newblock In {\em IEEE Conference on Computer Vision and Pattern Recognition
  (CVPR)}, June 2023.

\bibitem{yi2022generating}
Hongwei Yi, Hualin Liang, Yifei Liu, Qiong Cao, Yandong Wen, Timo Bolkart,
  Dacheng Tao, and Michael~J Black.
\newblock Generating holistic 3d human motion from speech.
\newblock In {\em IEEE Conference on Computer Vision and Pattern Recognition
  (CVPR)}, June 2023.

\bibitem{yu2021function4d}
Tao Yu, Zerong Zheng, Kaiwen Guo, Pengpeng Liu, Qionghai Dai, and Yebin Liu.
\newblock Function4d: Real-time human volumetric capture from very sparse
  consumer rgbd sensors.
\newblock In {\em Proceedings of the IEEE/CVF Conference on Computer Vision and
  Pattern Recognition}, pages 5746--5756, 2021.

\bibitem{zheng2022avatar}
Yufeng Zheng, Victoria~Fern{\'a}ndez Abrevaya, Marcel~C B{\"u}hler, Xu Chen,
  Michael~J Black, and Otmar Hilliges.
\newblock Im avatar: Implicit morphable head avatars from videos.
\newblock In {\em Proceedings of the IEEE/CVF Conference on Computer Vision and
  Pattern Recognition}, pages 13545--13555, 2022.

\bibitem{zheng2021pamir}
Zerong Zheng, Tao Yu, Yebin Liu, and Qionghai Dai.
\newblock Pamir: Parametric model-conditioned implicit representation for
  image-based human reconstruction.
\newblock {\em IEEE Transactions on Pattern Analysis and Machine Intelligence},
  pages 1--1, 2021.

\bibitem{Zheng2019DeepHuman}
Zerong Zheng, Tao Yu, Yixuan Wei, Qionghai Dai, and Yebin Liu.
\newblock Deephuman: 3d human reconstruction from a single image.
\newblock In {\em The IEEE International Conference on Computer Vision (ICCV)},
  October 2019.

\bibitem{zhu2022mvp}
Xiangyu Zhu, Tingting Liao, Jiangjing Lyu, Xiang Yan, Yunfeng Wang, Kan Guo,
  Qiong Cao, Stan~Z. Li, and Zhen Lei.
\newblock Mvp-human dataset for 3d human avatar reconstruction from
  unconstrained frames.
\newblock {\em arXiv preprint arXiv:2204.11184}, 2022.

\end{thebibliography}
}

\end{document}


\title{\ourtitle\\\textit{*Supplementary Material*}}


\author{ 
    Tingting Liao$^{1,2}$\thanks{Equal contribution.} \qquad
    Xiaomei Zhang$^{1,2 *}$  \qquad
    Yuliang Xiu$^{3}$ \qquad
    Hongwei Yi$^{3}$ \qquad
    Xudong Liu$^{4}$ \qquad \\ 
    Guo-Jun Qi$^{4,5}$ \qquad
    Yong Zhang$^{6}$ \qquad
    Xuan Wang$^{6}$ \qquad
    Xiangyu Zhu$^{1,2}$ \qquad
    Zhen Lei$^{1,2,7}$\thanks{Corresponding author.}   \\
    {\normalsize $^{1}$University of Chinese Academy of Sciences, Beijing, China ~~~}    \\
    {\normalsize$^{2}$MAIS, Institute of Automation, Chinese Academy of Sciences, Beijing, China} \\
    {\normalsize$^{3}$Max Planck Institute for Intelligent Systems, T\"ubingen, Germany ~~~} \\
    {\normalsize$^{4}$OPPO Research  ~~~ 
    $^{5}$Westlake University  ~~~ 
    $^{6}$Tencent AI Lab ~~~ 
    $^{7}$CAIR, HKISI, CAS }\\ 
    {\tt\small \{tingting.liao, xiaomei.zhang, xiangyu.zhu, zlei\}@nlpr.ia.ac.cn} \\  
     {\tt\small \{yuliang.xiu, hongwei.yi\}@tuebingen.mpg.de} \\  
    {\tt\small  \{yongzhang201303, xwang.cv, guojunq\}@gmail.com} \\  
    {\tt\small  \{xudong.liu\}@oppo.com} \\ 
}

\setcounter{figure}{7} 
\setcounter{table}{3} 
\setcounter{equation}{6} 

\maketitle

We provide more details for the method and experiments, as well as more quantitative and qualitative results, as an extension of \cref{sec: method}, \cref{sec: experiments} and \cref{sec: applications} of the main paper.

\section{Method \& Experiment Details}
\subsection{Dataset (\cref{sec: dataset})}

\qheading{Dataset size.}
We evaluate the performance of \modelname and \sota methods for a varying training-dataset size (\cref{fig:datasize,tab:geo-datasize}).
For this, we first combine \agora~ \cite{patel2021agora} ($3,109$ scans) and \thuman~\cite{zheng2019deephuman} ($600$ scans) to get $3,709$ scans in total. 
This new dataset is $8$x times larger than the $450$ \renderppl (``$450$-Rp'') scans used in \cite{saito2019pifu,saito2020pifuhd}. 
Then, we sample this ``$8$x dataset'' to create smaller variations, for $1/8$x, $1/4$x, $1/2$x, $1$x, and $8$x the size of ``$450$-Rp''. 

\smallskip
\qheading{Dataset splits.}
For the ``$8$x dataset'', we split the $3,109$ \agora scans into a new training set ($3,034$ scans), validation set ($25$ scans) and test set ($50$ scans). 
Among these, $1,847$ come from \renderppl~\cite{renderpeople} (see \cref{fig:kmeans-renderppl}), $622$ from \axyz~\cite{axyz}, $242$ from \humanalloy~\cite{humanalloy}, $398$ from \threeppl~\cite{3dpeople}, and we sample only $600$ scans from \thuman (see \cref{fig:kmeans-thuman}), due to its high pose repeatability and limited identity variants (see \cref{tab:dataset}), with the ``select-cluster'' scheme described below. 
These scans, as well as their \smplx fits, are rendered after every $10$ degrees rotation around the yaw axis, to totally generate {\tt $(3109 \text{~\agora} + 600 \text{~\thuman} + 150 \text{~\cape}) \times 36 = 138,924$} samples. 

\smallskip
\qheading{Dataset distribution via ``select-cluster'' scheme.} 
To create a training set with a rich pose distribution, we need to select scans from various datasets with poses different from \agora.
Following \smplify \cite{federica2016smplify}, we first fit a Gaussian Mixture Model (GMM) with $8$ components to all \agora poses, and \textbf{select} 2K \thuman scans with low likelihood. 
Then, we apply \mbox{M-Medoids} ({\tt $\text{n\_cluster} = 50$}) on these selections for \textbf{clustering}, and randomly pick $12$ scans per cluster, collecting $50 \times 12 = 600$ \thuman scans in total; see \cref{fig:kmeans-thuman}. 
This is also used to split \cape into \capeFP (\cref{fig:kmeans-cape-easy}) and \capeNFP (\cref{fig:kmeans-cape-hard}), corresponding to scans with poses similar (in-distribution poses) and dissimilar (\ood poses) to \agora ones, respectively.

\smallskip
\qheading{Perturbed \smpl.}
To perturb \smpl's pose and shape parameters, random noise is added to $\theta \text{ and } \beta$ by:
\begin{equation}
    \label{eq: perturb smpl}
    \begin{gathered}
        \theta \mathrel{+}= s_\theta * \mu  \text{,}    \\
        \beta  \mathrel{+}= s_\beta  * \mu  \text{,}    \\
    \end{gathered}
\end{equation}
where $\mu \in [-1,1]$, $s_\theta=0.15$ and $s_\beta=0.5$. 
These are set empirically to mimic the misalignment error typically caused by \offtheshell HPS during testing.

\highlight{
\smallskip
\qheading{Discussion on simulated data.}
The wide and loose clothing in \clothplus~\cite{bertiche2020cloth3d,madadi2021cloth3d++} demonstrates strong dynamics, which would complement commonly used datasets of commercial scans.
Yet, the domain gap between \clothplus and real images is still large. 
Moreover, it is unclear how to train an implicit function from multi-layer non-watertight meshes. Consequently, we leave it for future research.
}

\subsection{Refining \smpl (\cref{sec: body-refinement})}

\begin{figure}[h]
\centering
    \includegraphics[trim=006mm 002mm 006mm 002mm, clip=true, width=1.0 \linewidth]{photos/joint-optim-sanity.pdf}
    \caption{\smpl refinement error (y-axis) with different losses (see colors) and noise levels, $s_\theta$, of pose parameters (x-axis).}
    \label{fig:joint-optim-sanity}
\end{figure}

To statistically analyze the necessity of $\mathcal{L}_\text{N\_diff}$ and $\mathcal{L}_\text{S\_diff}$ in \cref{eq:body-fit}, we do a sanity check on \agora's validation set. 
Initialized with different pose noise, $s_\theta$ (\cref{eq: perturb smpl}), we optimize the $\{\theta, \beta, t\}$ parameters of the perturbed \smpl by minimizing the difference between rendered \smpl-body normal maps and \groundtruth clothed-body normal maps for 2K iterations. 
As \cref{fig:joint-optim-sanity} shows,  \textcolor{GreenColor}{$\mathcal{L}_\text{N\_diff} + \mathcal{L}_\text{S\_diff}$} always leads to the smallest error under any noise level, measured by the Chamfer distance between the optimized perturbed \smpl mesh and the \groundtruth \smpl mesh.

\begin{table*}
\centering
\resizebox{1.0\linewidth}{!}{
  \begin{tabular}{c|l|c|ccc|ccc|ccc|aaa}
    & Methods & \smplx & \multicolumn{3}{c|}{\agora-50} & \multicolumn{3}{c|}{\cape-FP} & \multicolumn{3}{c|}{\cape-NFP} & \multicolumn{3}{a}{\cape}\\
    & & condition. & Chamfer $\downarrow$ & P2S $\downarrow$ & Normal $\downarrow$ & Chamfer $\downarrow$ & P2S $\downarrow$ & Normal $\downarrow$ & Chamfer $\downarrow$ & P2S $\downarrow$ & Normal $\downarrow$ & Chamfer $\downarrow$ & P2S $\downarrow$ & Normal $\downarrow$\\
    \shline
     & $\text{\modelname}$ & \wcmark & 1.583 & 1.987 & 0.079 & \textbf{1.364} & \textbf{1.403} & 0.080 & \textbf{1.444} & \textbf{1.453} & 0.083 & \textbf{1.417} & \textbf{1.436} & 0.082\\
    \hline
    \multirow{4}{*}{D} & $\text{\smplx perturbed}$ & \wcmark & 1.984 & 2.471 & 0.098 & 1.488 & 1.531 & 0.095 & 1.493 & 1.534 & 0.098 & 1.491 & 1.533 & 0.097\\
    & $\text{\modelname}_\text{enc(I,N)}$ & \wcmark & 1.569 & \textbf{1.784} & \textbf{0.073} & 1.379 & 1.498 & \textbf{0.070} & 1.600 & 1.580 & \textbf{0.078} & 1.526 & 1.553 & \textbf{0.075}\\
    & $\text{\modelname}_\text{enc(N)}$ & \wcmark & \textbf{1.564} & 1.854 & 0.074 & 1.368 & 1.484 & 0.071 & 1.526 & 1.524 & \textbf{0.078} & 1.473 & 1.511 & 0.076\\
    & $\text{\modelname}_{\text{N}^{\dagger}}$ & \wcmark & 1.575 & 2.016 & 0.077 & 1.376 & 1.496 & 0.076 & 1.458 & 1.569 & 0.080 & 1.431 & 1.545 & 0.079
  \end{tabular}
}
\vspace{-0.5 em}
\caption{Quantitative errors (cm) for several \modelname variants conditioned on perturbed \smplx fits ($s_\theta = 0.15, s_\beta = 0.5$).}
\label{tab:sup-benchmark}
\end{table*}

\subsection{Perceptual study (\cref{tab: perceptual study})}

\qheading{Reconstruction on \inthewild images.}
We perform a perceptual study to evaluate the perceived realism of the reconstructed clothed \threeD humans from \inthewild images. 
\modelname is compared against $3$ methods, \pifu~\cite{saito2019pifu}, \pifuhd~\cite{saito2020pifuhd}, and \pamir~\cite{zheng2020pamir}.
We create a benchmark of $200$ unseen images downloaded from the internet, and apply all the methods on this test set.
All the reconstruction results are evaluated on \ac{amt}, where each participant is shown pairs of reconstructions from \modelname and one of the baselines, see \cref{fig:perceptual_study}. 
Each reconstruction result is rendered in four views: front, right, back and left. 
Participants are asked to choose the reconstructed \threeD shape that better represents the human in the given color image.
Each participant is given $100$ samples to evaluate.
To teach participants, and to filter out the ones that do not understand the task, we set up $1$ tutorial sample, followed by $10$ warm-up samples, and then the evaluation samples along with catch trial samples inserted every $10$ evaluation samples.
Each catch trial sample shows a color image along with either 
(1) the reconstruction of a baseline method for this image and the \groundtruth scan that was rendered to create this image, or 
(2) the reconstruction of a baseline method for this image and the reconstruction for a different image (false positive), see \cref{fig:perceptual_study:catch_trial}.
Only participants that pass $70\%$ out of $10$ catch trials are considered. 
This leads to $28$ valid participants out of $36$ ones. 
Results are reported in \cref{tab: perceptual study}.

\smallskip
\qheading{Normal map prediction.} 
To evaluate the effect of the body prior for normal map prediction on \inthewild images, we conduct a perceptual study against prediction without the body prior. 
We use \ac{amt}, and show participants a color image along with a pair of predicted normal maps from two methods. 
Participants are asked to pick the normal map that better represents the human in the image. 
Front- and back-side normal maps are evaluated separately. 
See \cref{fig:perceptual_study_normal} for some samples. 
We set up $2$ tutorial samples, 10 warm-up samples, $100$ evaluation samples and $10$ catch trials for each subject.
The catch trials lead to $20$ valid subjects out of $24$ participants. 
We report the statistical results in \cref{tab: suppl perceptual study norm}. 
A chi-squared test is performed with a null hypothesis that the body prior does not have any influence. 
We show some results in \cref{fig:normal_perceptual}, where all participants unanimously prefer one method over the other. 
While results of both methods look generally similar on front-side normal maps, using the body prior usually leads to better back-side normal maps.

\begin{table}[ht]
\centering
\footnotesize
\begin{tabular}{c|ccc}
 & w/ \smpl prior & w/o \smpl prior & P-value\\
\shline
Preference (front) & 47.3\% & 52.7\% & 8.77e-2\\
Preference (back) & 52.9\% & 47.1\% & 6.66e-2
\end{tabular}
\caption{Perceptual study on normal prediction.}
\label{tab: suppl perceptual study norm}
\end{table}

\begin{table}[t]
\centering
\footnotesize
\begin{tabular}{l|c|ccc}
& w/ global & pixel & point &  total \\
& encoder & dims & dims &  dims \\
\shline
\pifuSIM & \cmark & 12 & 1 & 13\\
\pamirSIM & \cmark & 6 & 7 & 13\\
$\text{\modelname}_\text{enc(I,N)}$ & \cmark &  6 & 7 & 13\\
$\text{\modelname}_\text{enc(N)}$ & \cmark & 6 & 7 & 13\\
$\text{\modelname}$ & \xmark & 0 & 7 & 7\\
\end{tabular}
\caption{Feature dimensions for various approaches. 
``pixel dims'' and ``point dims'' denote the feature dimensions encoded from pixels (image/normal maps) and \threeD body prior, respectively.}
\label{tab:feat_dim}
\end{table}

\subsection{Implementation details (\cref{sec: baseline models})}
    
\qheading{Network architecture.}
Our body-guided normal prediction network uses the same architecture as \pifuhd \cite{saito2020pifuhd}, originally proposed in \cite{justin2016perceploss}, and consisting of residual blocks with $4$ down-sampling layers.
The image encoder for \pifuSIM, \pamirSIM, and $\text{\modelname}_\text{enc}$ is a stacked hourglass  \cite{newell2016hourglass} with $2$ stacks, modified according to \cite{aaron2018volreg}. 
\cref{tab:feat_dim} lists feature dimensions for various methods; ``total dims'' is the neuron number for the first MLP layer (input). 
The number of neurons in each MLP layer is: $13$ ($7$ for \modelname), $512$, $256$, $128$, and $1$, with skip connections at the $3$rd, $4$th, and $5$th layers. 

\medskip
\qheading{Training details.}
For training $\normNet$ we do not use \thuman due to its low-quality texture (see \cref{tab:dataset}). On the contrary, $\imFunc$ is trained on both \agora and \thuman. The front-side and back-side normal prediction networks are trained individually with batch size of $12$ under the objective function defined in \cref{eq:normal-loss}, where we set $\lambda_\text{VGG} = 5.0$. 
We use the ADAM optimizer with a learning rate of $1.0 \times 10^{-4}$ until convergence at $80$ epochs. 

\newpage

\medskip
\qheading{Test-time details.}
During inference, to iteratively refine \smpl and the predicted clothed-body normal maps, we perform $50$ iterations \highlight{(each iteration takes $\sim460$ ms on a Quadro RTX $5000$ GPU)} and set $\lambda_\text{N} = 2.0$ in \cref{eq:body-fit}. 
\highlight{We conduct an experiment to show the influence of the number of iterations (\#iterations) on accuracy, see \cref{tab:loop}.}

The resolution of the queried occupancy space is $256^3$. 
We use \specific{rembg}\footnote{\url{https://github.com/danielgatis/rembg}} to segment the humans in \inthewild images, and use \specific{Kaolin}\footnote{\url{https://github.com/NVIDIAGameWorks/kaolin}} to compute per-point the signed distance, $\bodySDF$, and barycentric surface normal, $\bodyNormFeat$.

\begin{table}[h]
\footnotesize
\setlength{\parindent}{0em}
\addtolength{\tabcolsep}{-3pt}
\centering
\begin{tabular}{c|ccc}
    \multicolumn{1}{c|}{\# iters (460ms/it)} & 0 & 10 & 50\\
    \shline
    Chamfer $\downarrow$ & 1.417 & 1.413 & \textbf{1.339}\\
    P2S     $\downarrow$ & 1.436 & 1.515 & \textbf{1.378}\\
    Normal  $\downarrow$ & 0.082 & 0.077 & \textbf{0.074}\\
\end{tabular}
\caption{\modelname errors \wrt iterations}
\label{tab:loop}
\end{table}


\highlight{
\medskip
\qheading{Discussion on receptive field size.}
As \cref{tab:rp} shows, simply reducing the size of receptive field of \pamir does not lead to better performance. This shows that our informative 3D features as in \cref{eq:point feature} and normal maps $\predCloNormImg$ also play important roles for robust reconstruction. A more sophisticated design of smaller receptive field may lead to better performance and we would leave it for future research.
}

\begin{table}[h]
\footnotesize
\setlength{\parindent}{0em}
\addtolength{\tabcolsep}{-3pt}
\centering
\begin{tabular}{c|ccc}
    \multicolumn{1}{c|}{Receptive field} & 139 & 271 & 403\\
    \shline
    Chamfer $\downarrow$ & 1.418 & 1.478 & \textbf{1.366}\\
    P2S     $\downarrow$ & 1.236 & 1.320 & \textbf{1.214}\\
    Normal  $\downarrow$ & 0.083 & 0.084 & \textbf{0.078}\\
\end{tabular}
\caption{\pamir's receptive field}
\label{tab:rp}
\end{table}

\newpage

\section{More Quantitative Results (\cref{sec: evaluation})}
\Cref{tab:sup-benchmark} compares several \modelname variants conditioned on perturbed \smplx meshes. 
For the plot of \cref{fig:datasize} of the main paper (reconstruction error \wrt training-data size), extended quantitative results are shown in \cref{tab:geo-datasize}.

\begin{table}[h]
\centering{
\resizebox{\linewidth}{!}{
 \begin{tabular}{lc|ccccc}
    \multicolumn{2}{c|}{Training set scale} & 1/8x & 1/4x & 1/2x & 1x & 8x\\
    \shline
    \multirow{2}{*}{\pifuSIM } & Chamfer $\downarrow$ & 3.339 & 2.968 & 2.932 & 2.682 & 1.760 \\
    & P2S $\downarrow$ & 3.280 & 2.859 & 2.812 & 2.658 & 1.547 \\
    \multirow{2}{*}{\pamirSIM } & Chamfer $\downarrow$ & 2.024 & 1.780 & 1.479 & 1.350 & \colorbox{LightCyan}{1.095} \\
    & P2S $\downarrow$ & 1.791 & 1.778 & 1.662 & 1.283 & 1.131 \\
    \multirow{2}{*}{\modelname} & Chamfer $\downarrow$ & 1.336 & 1.266 & 1.219 & \colorbox{LightCyan}{1.142} & \colorbox{LightCyan}{\textbf{1.036}}\\
    & P2S $\downarrow$ & 1.286 & 1.235 & 1.184 & \colorbox{LightCyan}{1.065} & \colorbox{LightCyan}{\textbf{1.063}}\\
 \end{tabular}}}
\vspace{-0.5 em}
\caption{Reconstruction error (cm) \wrt training-data size. 
``Training set scale'' is defined as the ratio \wrt the $450$ scans used in \cite{saito2019pifu,saito2020pifuhd}. 
The ``8x'' setting is all $3,709$ scans of \agora~\cite{patel2021agora} and \thuman~\cite{zheng2019deephuman}.  
\colorbox{LightCyan}{Results} outperform \groundtruth \smplx, which has 1.158 cm and 1.125 cm for Chamfer and P2S in \cref{tab:benchmark}.}
\label{tab:geo-datasize}
\end{table}

\section{More Qualitative Results (\cref{sec: applications})}
\Cref{fig:comparison1,fig:comparison2,fig:comparison3} show reconstructions for \inthewild images, rendered from four different view points; normals are color coded.
\Cref{fig:out-of-frame} shows reconstructions for images with \oof cropping.
\Cref{fig:failure cases} shows additional representative failures.
The \video on our website shows animation examples created with \modelname and \scanimate. 

\begin{figure*}[t]
  \centering
  \vspace{-3.5 em}
  \begin{subfigure}{0.32\linewidth}
    \includegraphics[width=\linewidth,frame]{photos/agora.png}
    \caption{\renderppl~\cite{renderpeople} (450 scans)}
    \label{fig:kmeans-renderppl}
  \end{subfigure}
  \begin{subfigure}{0.64\linewidth}
    \includegraphics[width=\linewidth,frame]{photos/thuman.png}    
    \caption{\thuman~\cite{zheng2019deephuman} (600 scans)}
    \label{fig:kmeans-thuman}
  \end{subfigure}
  \begin{subfigure}{0.32\linewidth}
    \includegraphics[width=\linewidth,frame]{photos/cape-easy.png}
    \caption{\capeFP~\cite{ma2020cape} (fashion poses, 50 scans)}
    \label{fig:kmeans-cape-easy}
  \end{subfigure}
  \begin{subfigure}{0.64\linewidth}
    \includegraphics[width=\linewidth,frame]{photos/cape-hard.png}
    \caption{\capeNFP~\cite{ma2020cape} (non fashion poses, 100 scans)}
    \label{fig:kmeans-cape-hard}
  \end{subfigure}
  \caption{Representative poses for different datasets.}
  \label{fig:kmeans}
\end{figure*}

\begin{figure*}
     \centering
    \begin{subfigure}{0.49\linewidth}
        \includegraphics[width=\linewidth]{photos/geometry_tutorial.png}
        \caption{A tutorial sample.}
    \label{fig:perceptual_study:tutorial}
    \end{subfigure}
    \hfill
    \begin{subfigure}{0.49\linewidth}
        \includegraphics[width=\linewidth]{photos/0063.png}    
        \caption{An evaluation sample.}
    \label{fig:perceptual_study:evaluation}
    \end{subfigure}
    \centering
    \begin{subfigure}{\linewidth}
        \includegraphics[width=0.48\linewidth]{photos/000_top.png}
        \hfill
        \includegraphics[width=0.48\linewidth]{photos/008_top.png} 
        \caption{   Two samples of catch trials. 
                    Left:  result from this image (top) vs from another image (bottom). 
                    Right: \groundtruth (top) vs reconstruction mesh (bottom).}
        \label{fig:perceptual_study:catch_trial}
    \end{subfigure}
    
    \caption{Some samples in the perceptual study to evaluate \textbf{reconstructions} on \inthewild images.}
    \label{fig:perceptual_study}
\end{figure*}
\begin{figure*}[t]
    \centering
    \begin{subfigure}{\linewidth}
        \includegraphics[width=0.5\linewidth]{photos/norm_tutorial1.png}
        \includegraphics[width=0.5\linewidth]{photos/norm_tutorial2.png}
        \caption{The two tutorial samples.}
    \label{fig:perceptual_study_normal:tutorial}
    \end{subfigure}
    \begin{subfigure}{0.49\linewidth}
        \includegraphics[width=0.49\linewidth]{photos/062.png} 
        \hfill
        \includegraphics[width=0.49\linewidth]{photos/347.png} 
        \caption{Two evaluation samples.}
    \label{fig:perceptual_study_normal:evaluation}
    \end{subfigure}
    \hfill
    \begin{subfigure}{0.49\linewidth}
        \includegraphics[width=0.49\linewidth]{photos/003_top.png}
        \hfill
        \includegraphics[width=0.49\linewidth]{photos/009_bottom.png} 
        \caption{Two catch trial samples.}
    \label{fig:perceptual_study_normal:catch_trial}
    \end{subfigure}
    \caption{Some samples in the perceptual study to evaluate the effect of the \textbf{body prior} for \textbf{normal prediction} on \inthewild images.}
    \label{fig:perceptual_study_normal}
\end{figure*}

\begin{figure*}[t]
    \centering
    \begin{subfigure}{\linewidth}
    \includegraphics[trim=000mm 000mm 000mm 000mm, clip=True, width=1.00 \linewidth]{photos/normal-random_front.pdf}
    \caption{Examples of perceptual preference on \textbf{front} normal maps. Unanimously preferred results are in \Ovalbox{black boxes}. The back normal maps are for reference.\\}
    \vspace{-0.5em}
    \label{fig:normal_perceptual:front}
    \end{subfigure}
    \begin{subfigure}{\linewidth}
    \includegraphics[width=1.00 \linewidth]{photos/normal-random_back.pdf}    
    \caption{Examples of perceptual preference on \textbf{back} normal maps. Unanimously preferred results are in \Ovalbox{black boxes}. The front normal maps are for reference.}
    \vspace{-0.5em}
    \label{fig:normal_perceptual:back}
  \end{subfigure}
  \caption{Qualitative results to evaluate the effect of body prior for normal prediction on \inthewild images.}
  \vspace{-0.5em}
  \label{fig:normal_perceptual}
\end{figure*}

\newcommand{\comparisonCaptionSupMat}{Qualitative comparison of reconstruction for \textcolor{GreenColor}{\modelname} vs \sota. Four view points are shown per result.}

\begin{figure*}
    \centering
    \vspace{-2.0 em}
    \hspace{-3.0 em}
    \includegraphics[width=1.07 \textwidth]{photos/compare1.pdf}
    \caption{\comparisonCaptionSupMat}
    \label{fig:comparison1}
\end{figure*}

\begin{figure*}
    \centering
    \vspace{-2.0 em}
    \hspace{-3.0 em}
    \includegraphics[width=1.07 \textwidth]{photos/compare2.pdf}
    \caption{\comparisonCaptionSupMat}
    \label{fig:comparison2}
\end{figure*}

\begin{figure*}
    \centering
    \vspace{-2.0 em}
    \hspace{-3.0 em}
    \includegraphics[width=1.07 \textwidth]{photos/compare3.pdf}
    \caption{\comparisonCaptionSupMat}
    \label{fig:comparison3}
\end{figure*}

\begin{figure*}
    \centering
    \includegraphics[width=1.04 \linewidth]{photos/out-of-frame.pdf}
    \caption{Qualitative comparison (\textcolor{GreenColor}{\modelname} vs \sota) on images with out-of-frame cropping.}
    \label{fig:out-of-frame}
\end{figure*}

\begin{figure*}
\centering
    \includegraphics[width=1.04 \linewidth]{photos/failure_cases.pdf}
    \caption{More failure cases of \modelname.}
    \label{fig:failure cases}
\end{figure*}

\clearpage
{\small
\balance
\bibliographystyle{config/ieee_fullname}
\bibliography{config/BIB}
}